\documentclass[a4paper]{article}
\pdfoutput=1
\usepackage[utf8]{inputenc} 
\usepackage[T1]{fontenc}    
\usepackage{hyperref}       
\usepackage{url}            
\usepackage{booktabs}       
\usepackage{amsfonts}       
\usepackage{nicefrac}       
\usepackage{microtype}      
\usepackage{amsmath,amssymb} 
\usepackage{color}
\usepackage{algorithmic}
\usepackage{algorithm}
\usepackage{enumitem}
\usepackage{tikz}
\usepackage{authblk}
\usepackage{wrapfig}
\usepackage{mathtools}
\usetikzlibrary{bayesnet}

\def\be{\begin{equation}}
\def\ee{\end{equation}}
\def\~{\mathaccent "7E}

\def\ee{\mathbf{e}}

\def\ee{\mathbf{e}}

\newcommand{\NNN}{{\mathcal{N}}}

\def\00{\mathbf{0}}
\def\11{\mathbf{1}}

\newcommand{\off}[1]{}

\def\eg{e.g.,\ }
\def\etal{et\ al.\ }

\newcommand{\jwf}[1]{{{\color{green}}}} 
\newcommand{\gr}[1]{{{\color{blue}}}} 

\DeclarePairedDelimiter{\round}{\lfloor}{\rceil}

\graphicspath{{figures/}}
\newcommand\gpsFigureWidth{0.20}
\newcommand\figurefontsize{\footnotesize}

\title{A Nonparametric Model for Multimodal Collaborative Activities Summarization}

\author{Guy Rosman}
\author{John W. Fisher III}
\author{Daniela Rus}
\affil{CSAIL, MIT\\
  32 Vassar St, Cambridge, MA 02139\\
  \texttt{\{rosman|fisher|rus\}@csail.mit.edu} 
}

\begin{document}

\maketitle

\begin{abstract}
\jwf{
\begin{itemize}
\item ego-centric data provides a \emph{unique} opportunity to pool
  data and reason about joint behaviors
\item rich multi-modal \emph{ego-centric} data streams provide...
\item ego-centric data sources $\longrightarrow$ collaborative
  activity analysis
\end{itemize}
}
Ego-centric data streams provide a \emph{unique} opportunity to reason
about joint behavior by pooling data across individuals. This is
especially evident in urban environments teeming with human
activities, but which suffer from incomplete and noisy data.
Collaborative human activities exhibit common spatial, temporal, and
visual characteristics facilitating inference across individuals from
multiple sensory modalities as we explore in this paper from the
perspective of meetings.

We propose a new Bayesian nonparametric model that enables us to  efficiently  pool video and GPS data towards collaborative activities analysis from
multiple individuals. We demonstrate the utility of this model for inference tasks
such as activity detection, classification, and summarization.
We further demonstrate how spatio-temporal structure embedded in our model enables better understanding of partial and noisy observations 
such as localization and face detections based on social interactions.  
We show results on both synthetic experiments and a new dataset of
egocentric video and noisy GPS data from multiple individuals.

\end{abstract}

\section{Introduction}
  
\gr{Add ATUS categories examples, maybe non-hierarchical GANTT chart?}
Consider individuals equipped with wearable cameras and GPS devices
going about an urban center, meeting and interacting with each
other. Aggregation of their \emph{ego-centric} data streams leads to a
richer understanding of their collaborative experience, whereas
analysis of the individual streams suffers from noisy, partially denied, and unstructured signals.
%
%
Here, we suggest an approach to infer collaborative activities such as
``meetings'' or ``interactions'' from streaming wearable sensors based
on a collective multi-modal generative model.  Inference allows us to
summarize and perform analysis in the context of life-logging,
detecting special events from multiple sensors, and supporting
coordinated human activities. Interestingly, by combining multiple
\emph{ego-centric} data streams, one can infer the activities of one
individual using the data of another. The underlying model handles
noisy and partially denied data streamed from multiple (wearable)
devices, a crucial capability in noisy sensory environments.

By treating collaborative activity instances as latent factors with a
spatio-temporal structure, we can summarize activities for co-located
individuals despite noisy and partial/missing signals by pooling data
for the purpose of collaborative reasoning.  Human observers can
understand such content (at small scale) with the appropriate
visualizations, partially because we understand the spatio-temporal
and semantic structure of human activities.
We seek to
automate such abilities via an inference procedure that summarizes collective and individual experiences via the aggregated
data streams.


We explore meetings between people as a case study for detecting,
characterizing, and utilizing the temporal and collaborative structure of activities. 
We examine summarization as a task that benefits from higher-level information, 
even in the cases that activity detections are uncertain given the observations. Pooled information from
multiple sensors (\eg video, GPS) and multiple individuals, coupled with a
structured model for activities, allows us to handle partial
information in both GPS and video data and reduce the uncertainty regarding the participants across modalities. 
The proposed approach is
robust to missing data and low-quality, poor-vantage point footage
that is unsuitable for strong geometric reasoning such as 3D
reconstruction. Consider, for instance, a commute via a
crowded subway or a family ski trip. Such scenarios include non-rigid
3D motion, missing and noisy GPS observations \cite{NAV:8292634},
abrupt viewpoint changes, and video segments which lack features or
textures.  Yet, a casual observer of videos and GPS traces from these
activities could readily piece together a common story without the
need for a full model of the environment. 


\textbf{Contributions} Here we present a model that captures these characteristics.
\begin{enumerate}[topsep=0pt,itemsep=-1ex,partopsep=1ex,parsep=1ex,label=\roman*]
 \item We propose a probabilistic model and inference algorithms for the explanation
of egocentric video and GPS data in terms of human activities.  This model allows us to
handle multiple, overlapping activity definitions and partial data in all
modalities, by pooling data from across individuals and modalities.
\item Using our model, we demonstrate detection and summarization of
  \emph{collaborative} human activities from egocentric cameras and GPS signals
  in an urban environment where both the visual and the GPS signals are either
  missing or distorted.
\item We demonstrate improved localization of individuals conditioned on detected activities from corrupt and partial sources such as GPS and egocentric video streams.
\item We provide a dataset to allow further annotation and examination of similar methods, in order to foster future discussion and improvements.
\end{enumerate}
Our model is derived from simple assumptions, and we show its
relevance to complex tasks on multimodal data. 
Activities are detected through a sampling approach for the activity instances and their parameters. 
The model is easily extensible to other sensory modalities, such as inertial sensors \cite{bulling2014tutorial}, or textual sources \cite{weng2011event}.
We describe the
assumptions and the model in Section~\ref{sec:model}, and the associated activity inference, localization, and summarization
procedures in Section~\ref{sec:inference}.  In Section~\ref{sec:results} we demonstrate our ability to
infer the activities and leverage this analysis towards improved localization and understanding of people's story 
as portrayed by the data.  Section~\ref{sec:conclusions} discusses
possible extensions and concludes the report. In Appendix~A we present extensions to the social dynamic models described below. This is followed by additional details on the sampling procedure in Appendix~B and a approach for video summarization based on the sampled activity instances in Appendix~C.


\begin{figure}
    \includegraphics[width=\linewidth]{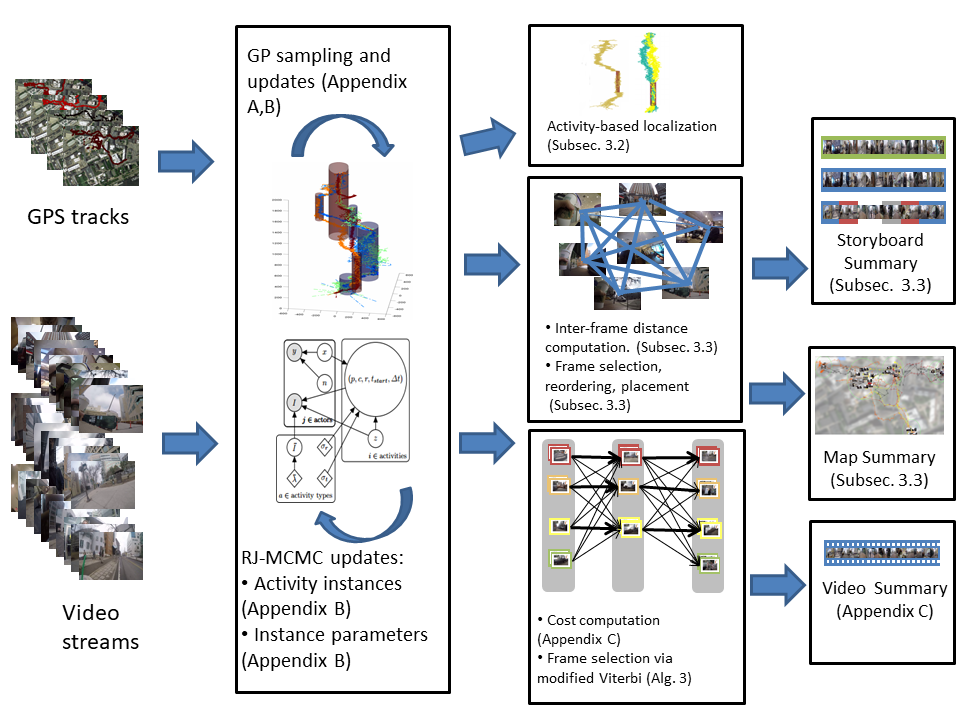} 
  \caption{\figurefontsize Our model reasons over multiple video and 
    GPS streams, inferring and summarizing collaborative interactions. We employ a generative model utilizing Gaussian processes for GPS and learned likelihood models for the semantic visual content and recognized faces.
    This enables summarization of the participants experience, detection of specific activities, and correction of the individuals' partial GPS and visual detections streams. \label{fig:overview} }
\end{figure}

\subsection{Related work}
\begin{figure}
\centering
\includegraphics[width=1\linewidth]{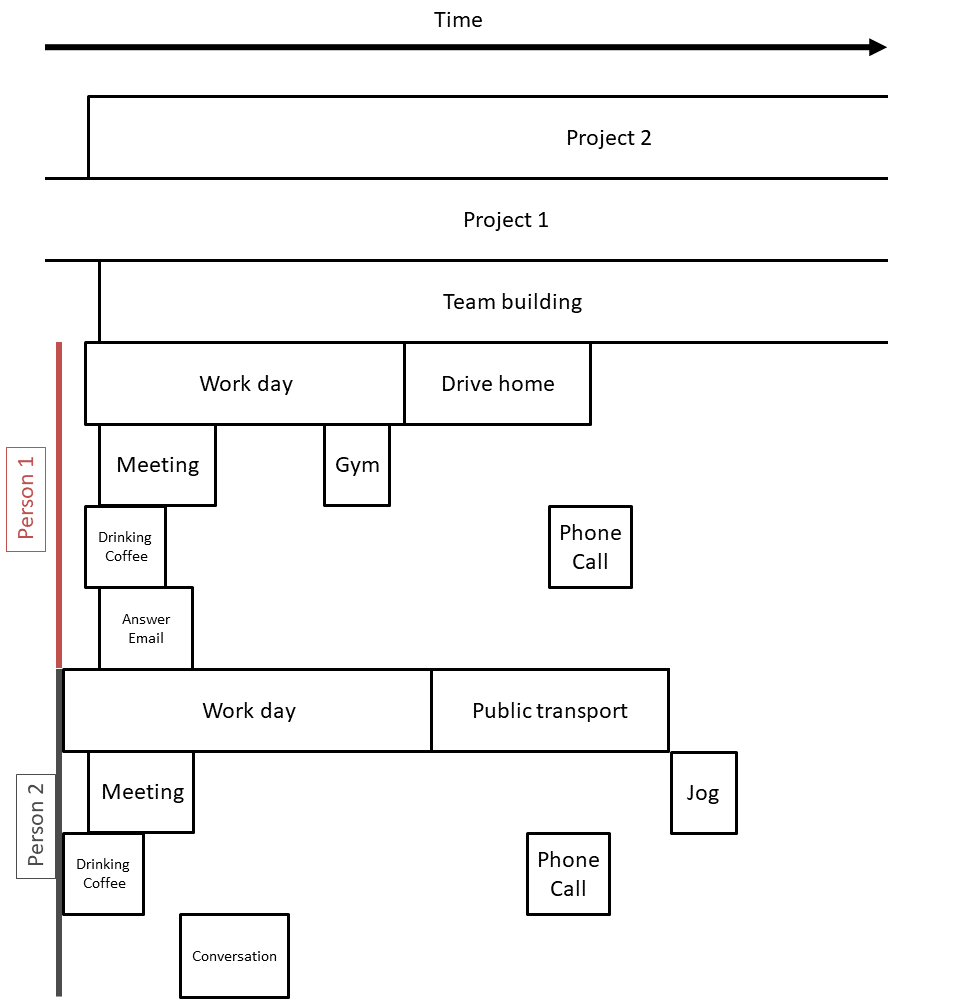} 

 \caption{\label{fig_time_intervals}Time intervals for several activities in a work/office scenario involving 2 people. Note the non-hierarchical structure formed between activities even for a single person, let alone multiple participants. Some activities are best inferred via visual data (e.g. drinking coffee) while others are easily inferred via GPS (\eg driving home), or require significant priors on human activities to infer (\eg team building / project 1).
 Some activities in this figure correspond to American Time-Use Survey(ATUS)\cite{shelley2005developing} Lexicon category codes such as 180501 (drive home / public transport), 050201 (phone call), 050202 (drinking coffee), 130124 (jogging), while others are higher-level (team building, project 1) or potentially lower-level.}
 
\end{figure}

Our paper relates to several lines of research.
These have attempted to tackle activity inference using GPS,
visual, and other sensor data (for example,
\cite{Liao:2007:EPA:1229555.1229562},
\cite{Yuan:2008:MGT:1460096.1460099,caba2015activitynet,yan2015egocentric}
and \cite{Huynh:2008:DAP:1409635.1409638}, respectively). 
However, our model reasons over latent 
locations and activities \emph{conditioned} on the set of visual and GPS cues,
marginalizing over possible interpretations of the data.  
We do not assume that
activities form partitions or hierarchies over the timeline except in
specific cases, even for a single person.  This is due both to a
non-trivial overlap structure and to multiple, equally-plausible
interpretation of the observations.  Such structures lend themselves
to probabilistic models that can describe multiple interpretations.
For an illustration of the possible relation between different activities' time span, consider the illustrative example in Figure~\ref{fig_time_intervals}.

Sampling allows us to reason about activities
without committing to a single ``correct'' explanation.  This approach
of multiple explanations can be maintained, as we show, throughout the
analysis while gracefully handling partial/corrupt data.

Several lines of research have explored
multiple video streams for reasoning over collaborative
activities. These methods generally assume that all
participants observe the same scene with uninterrupted data streams
\cite{6197723,Fu:2010:MVS:2219116.2219978},
\cite{DBLP:conf/cvpr/HoshenBP14,DBLP:journals/corr/Fu14}, or that the
3D environment can be reliably inferred
\cite{DBLP:conf/iccv/ParkJS13,Arev:2014:AEF:2601097.2601198,joo_iccv_2015}.
The proposed method does not rely on such restrictive assumptions,
allowing for partial, discontinuous, and noisy observations over
multiple scenes amongst varying subsets of participants and eschewing
reconstruction as an intermediate step altogether. The method is
robust to non-rigid 3D motion and fast rotations, as well as GPS
denial.  Several efforts have been dedicated to meaningfully choosing which
data to display and when, regarding both frame selection
\cite{Fu:2010:MVS:2219116.2219978}, and alignment
\cite{6197723,DBLP:conf/cvpr/DaleSAP12}.  Quality measures for each
shot are often used in a constrained graph traversal
\cite{Zhang:2015:ASM:2716635.2659520,Arev:2014:AEF:2601097.2601198}.
There, activity inference is treated separately from the
understanding of how different sources relate to one another.  While
incorporation of semantics for summarization was done, for example, by
Tat-Jen \etal \cite{TatJen06} and Lu \etal
\cite{Lu:2013:SSE:2514950.2516026,lee2015predicting}, it was done so
only at the feature level of objects/instantaneous
interaction, rather than over complete activities, due to a variety of
challenges including the difficulty of classifying the start and end of
activities in an arbitrary video, be it egocentric or third-person.
Inferring human activities from such data is non-trivial, particularly
in the absence of reliable activity labelling.




\section{Model}
\label{sec:model}
We define a \emph{collaborative} activity as one in which two or more
individuals, or \emph{actors} (as we call them henceforth), are
spatially proximate for some finite duration accompanied by
distinctive categorical features.  \jwf{I think we should state that
  we don't define these explicitly, but that we learn both features
  and spatio-temporal extent.}  While we emphasize collaborative
summarization and joint activities, the model may encompass individual
activities as well. A key attribute of the model is that by
integrating multiple \emph{ego-centric} data streams, one can infer
the activities of one actor using the data of a different actor.

We define activities as having a
categorical \emph{activity type} carrying some semantic meaning --- in our data and prior feature distributions we exemplify ``coffeeshop''
vs. ``chance street meeting''. We could extend to other fixed or inferred types, \eg ``work meeting'' or ``gym workout''. These types parameterize the prior
distribution of meeting characteristics, such as spatio-temporal
extent and visual content. \jwf{As I've stated in the past, I think
  providing specific \emph{semantic} descriptions of activities is
  problematic since we don't actually do that in the paper. The risk
  is that reviewers will think we are overstating our claims.}
  \gr{I narrowed it down to say what activities are from our data and
    features, and what activities could easily be in the model, given
    a different set of features and data.}
\jwf{I am referring to the use of things like ``chance street
  meeting'' above and ``gym-practice'' below. We have no such
  \emph{semantic} labels, only \emph{categorical} labels. We need to
  be clear when we use such terminology that it is for making an
  analogy rather than implying that we have explicit semantic types.}
For each activity type there may be multiple instances, or
\emph{activities} of that type. The $i$-th activity instance has a
static center $c_i$ and radius $r_i$.  A \emph{configuration} is the
set of all activity instances occurring and explaining the
observation. Subsection~\ref{subsec:infinite_model} describes a
probabilistic model that defines the probability of a set of
observations given a configuration and its parameters. Subsequently,
configurations may be sampled from this model.

Activity instances of different types may overlap in time, space, and
participants depending on the activity types involved, e.g lunch may
be part of a working day, but sleep and gym-practice do not
overlap.  
The activity set of each actor is far from being a partition, as
assumed in many models for activity recognition.  Instead, each
activity is an uncertain function of the physical/geometric scene
content and participants. This view favors a
nonparametric model where activity instances are random occurrences.

Here, we discuss the mathematical model by which we represent the
trajectories of actors, their collaborative activities, and related
visual cues. The overlap structure of different activity types is assumed to be known in advance as part of the model definition (\eg whether two activity type can overlap or not, or if there are containment relations between activities).

\vspace{-0.25cm}
\subsection{Modeling Participant Location}
\vspace{-0.25cm}
\label{subsec_gp_latent}
%
We use a latent Gaussian process $x$, inferred via noisy GPS observations
$y$ to represent the location and trajectory of each participant. 
\begin{flalign}
\label{eq:xy_gp}
x \sim \mathcal{GP}(\textit{0},k), \ \ \ y=x+n, n \sim \NNN(0,\Sigma)
\end{flalign}
%


%
This representation facilitates incorporating multiple data streams
with differing sampling rates and missing measurements since activity
\emph{reasoning} is performed over the \emph{latent} trajectories. We
experimented with both Mat\'{e}rn-class kernels and inverse-linear
kernels $k(t,s)=\frac{1}{\|t-s\|}$, with the latter yielding better
performance in practice, though we note that the former has several
advantages including integrability, differentiability and controlled
smoothness \cite{stein1999}. Examples of measurements and estimated
paths are shown in Figure~\ref{fig:localization}.

\vspace{-0.25cm}
\subsection{A Nonparametric Model for Human Activities}
\vspace{-0.25cm}
\label{subsec:infinite_model}
We formulate activity detection as a Bayesian nonparametric inference
problem where (possibly infinite) activity instances are created to explain the data, along with inferred
parameters. Let $A_i$ denote the $i$-th activity instance while
$\{A_i\}_{i=1}^{n}$ denotes the set of all activities (more properly a
realization of such), whose number and parameters are the focus of inference. 
\begin{figure}
\centering
\includegraphics[width=\linewidth]{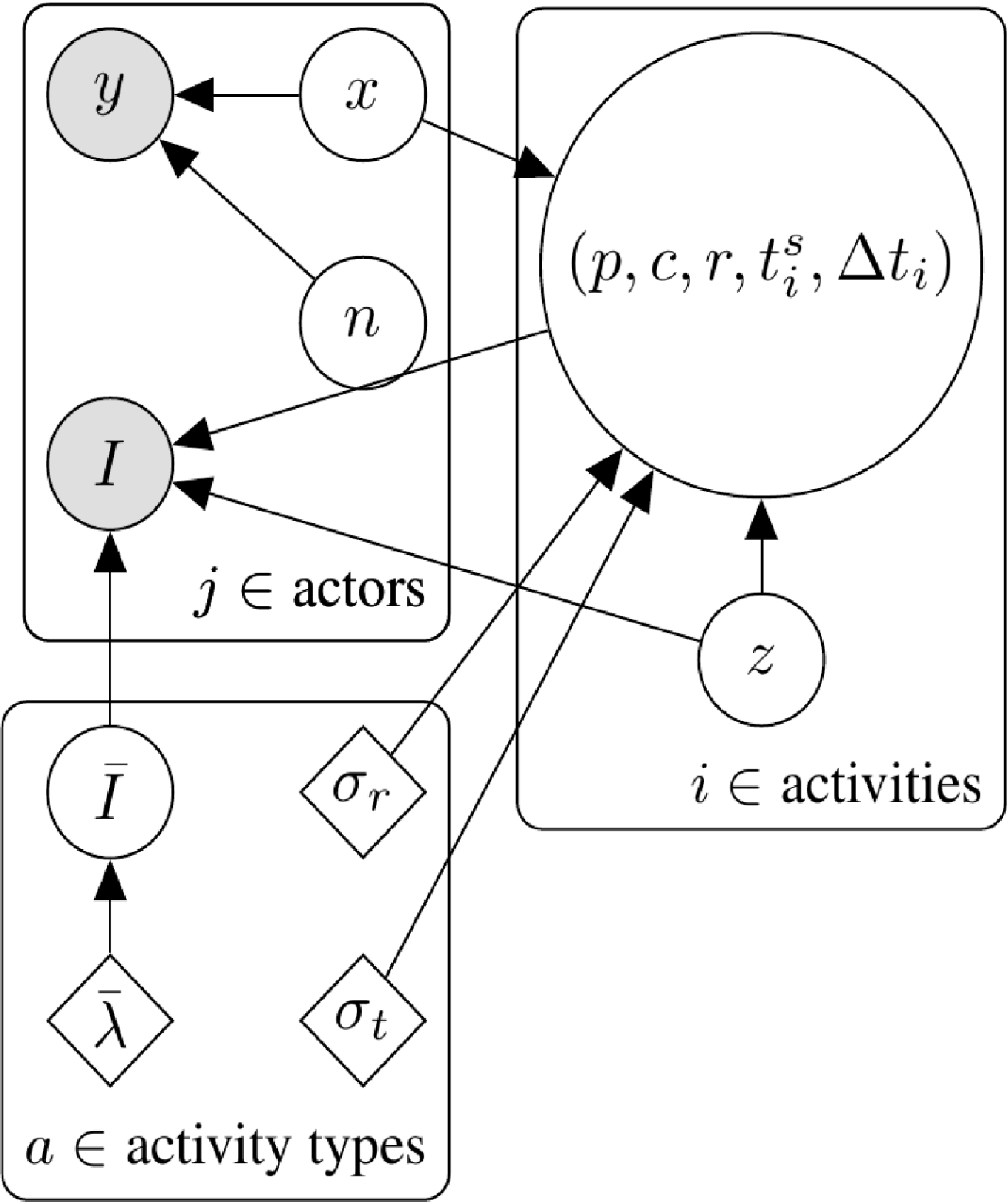}
\caption{\figurefontsize Proposed model for activity instance inference. Plate notation notes the separate instances of actors' latent and observed data, activity types (here assumed to be learned in advance), and activity instances. \label{fig_activities_pgm}} 
\end{figure}
Figure~\ref{fig_activities_pgm} depicts the graphical model used for
inference. The associated probability model
for explains the observations
$y,I$, as well as a latent actor trajectory $x$ in terms of the activity configuration $\{A_i\}$, and is given by \jwf{It seems odd to express it this way since $x$
  is a latent variable. }\gr{I reworded, and added a note saying that we marginalize over $x$. I think it's clearer with just a sentence and not an equation, but I am okay both ways.}
{\small\begin{flalign}
\label{eq:complete_prob}
   p(x,y,I|\{A_i\})\propto &  \varphi_{\textrm{GP}}(x,y|\{t_i,\Delta t_i,p\})\varphi_{\textrm{C}}(\{A_i\}|x) \times \\
 \nonumber &\prod_i \varphi_{\textrm{SC}}(I|\bar{I},z_i)\varphi_{\textrm{FC}}(I|\bar{I},z_i) \varphi_{\textrm{R}}(r_i)\varphi_{\textrm{SP}}(\Delta t_{i}) \times 
 \varphi_{\textrm{PR}}\left(A_i|x,p,c,r,t_i^s,\Delta t_i\right),
\end{flalign}}
where $\varphi_{\textrm{GP}}$ is the distribution of the Gaussian conditioned samples from Subsection~\ref{subsec_gp_latent}. $\varphi_{C},\varphi_{SC},\varphi_{R},\varphi_{SP},\varphi_{PR}$ are
factors that relate to data coverage and visual features, activity span priors and the probability for having activities. These are detailed in the remainder of the section, and specifically, Equations~\ref{eq:xy_gp},
\ref{eq:coverage}, \ref{eq:pres}. While specific modeling choices are
made here, modifications are conceptually straightforward. During our inference we sample for $\{A_i\}$ while marginalizing over $x$.

%
%
%

We penalize the \emph{lack} of association of trajectories to an event
in order to avoid the trivial result that \emph{no} activities occur
as follows,
\begin{align}
\label{eq:coverage}
 \varphi_{\textrm{C}}\left(\{A_i\}|\{x\}\right) \propto \exp\left\{-c_{u}\sum_{t_i \in obs(p)} I_{\text{unexp}}(p,t_i) \right\},
\end{align}
where $I_{\text{unexp}}(p,t_i)$ is an indicator that observation $t_i$ of actor $p$ was not covered by any activity in $\{A_i\}$. 
This depends on the latent trajectory $x_p$, and the spatio-temporal
span of each $A_i$. We set the constant $c_{u}$ empirically.

Each activity has a \textbf{spatio-temporal extent}  with associated start time
$t_i^{s}$ (with a uniform prior)  and time span $\Delta
t_i$. 
The time span is distributed as log-normal, represented by $\varphi_{\textrm{SP}}(\Delta t_i;\mu_t,\sigma_t)$.
Each activity is centered around a point in space where the radius of the
activity is also distributed as log-normal, denoted by $\varphi_{\textrm{R}}(r_i,\mu_r,\sigma_{r,t})$.

Each activity has a set of associated \textbf{participants} where the
number of participants is drawn from a discretized log-normal
distribution and the identities of the participants is drawn uniformly
from the set of actors.
We assume that participants are present in the spatial extent for the duration of the activity, centered around point $c_i$ (uniformly sampled). This is captured by a factor
\begin{align}
\label{eq:pres}
\varphi_{\textrm{PR}}\left(A_i|\begin{matrix}
\{x\},p,c,r,\\
t_i^s,\Delta t_i                           
                          \end{matrix}
\right) \propto \exp\left\{-\int h_{r_i}^{c_i}(x,t)dt\right\},
\end{align}
where $h_{r_i}^{c_i}$ is an indicator function limiting excursion of trajectory $x$ to a maximum distance $r_i$ from the center $c_i$ at time $t$.

We incorporate scene classification features as \textbf{visual cues}
for each frame using the Places CNN model \cite{NIPS2014_5349}.
We associate frames with each activity according to whether or not
their location and time is within the span of the activity.  The
probability of each scene classifier is modeled as a normal, separable
distribution in both features and frames,
$\varphi_{\textrm{SC}}(I|\bar{I},z) \propto \prod_v N(\mu_{v,z},
\sigma_{v,z}^2)$.
$\mu_{v,z}$ and $\sigma_{v,z}^2$ represent the mean and variance
of each meeting type's visual content, per the Places feature, and $z$
represents the activity type.
These were estimated based on a
set of photos taken at the same locations as the experiments. \jwf{what does this mean? Do
we have specific locations or just types of locations via Places?}\gr{We took Places features in the same locations as the experiments and computed the (feature-wise) distribution over these 2 sets of feature vectors. I added a few words to make it clearer.}

Face detections are also used as a visual cues. We use
face frontalization \cite{DBLP:conf/cvpr/HassnerHPE15} and one-shot
similarity kernels \cite{DBLP:conf/iccv/WolfHT09}, along with a set of
precollected and annotated face examples to detect and recognize
people in the video, thus 
creating a set of detection streams.  Given an
activity and its participants, we treat the detections of each actor
as counts from a Poisson process with either a high rate
$\lambda^{p}_{p}$ if the actor $p$ is a participant, or a low rate
$\lambda^{n}_{p}$ if the actor is not a participant in the activity.

\subsection{Conditioning Trajectories on Activities}
\label{subsec_conditioning_gp}
For a given activity instance, the participants are more likely to be at the
same location, depending on activity type.
While this is captured by the
spatial factor in Equation~\ref{eq:pres}, we may have impoverished samples \eg due to GPS denial. 
In such cases, the trajectories of \emph{all} participants traveling
outside the spatial extent of the activity should be \emph{rare}
event and should be sampled judiciously. While rare events in Gaussian random fields have been studied
extensively, the majority of the work has been on excursion sets which are less
applicable in our case (see \cite{adler2000,KratzLeon10} and references
therein). 

We accommodate this concern in the following manner. Given a
sampled configuration, we condition on auxiliary observations of the proximity of the
participants.  For spatially static activities such as the ones we show in Section~\ref{sec:results},  the auxiliary observations have the form
\begin{align*}
 0=x_p(t) - \frac{1}{|P_a| T} \sum x_{p'}(t) + n_{(p,a)}(t).
\end{align*}
A different set of auxiliary observations, suitable for dynamic group activities is shown in Appendix~A. Once we have added the auxiliary observations, we condition on all observations to obtain samples of the latent trajectories (which are now statistically dependent). 
This can be seen in Figure~\ref{fig:localization}(g,h), when determining the conditioned trajectories, but we note that
for the scenarios we show in Section~\ref{sec:results}, the GPS denial
periods were sufficiently short to inference the activities regardless of this problem. \jwf{Do we
  really need to call attention to this if our experiments don't
  really have to deal with it? It is an important point, but seems to
  be muddying the message.}\gr{I revised to emphasize that we do see the correction in the trajectories. I think it's a phenomenon that people tend to forget/ignore so it's important to highlight, even if we can attach it to just one figure.. but I'm ok with dropping.}


\section{Activity Summarization and Analysis}
\label{sec:inference}
We now describe the inference procedures relevant for our activity model, and our use of it for summarization and localization.

\vspace{-0.15cm}
\subsection{Activity Inference}
\vspace{-0.15cm}
As noted previously, reasoning over activities utilizes inferred
trajectories, in our case samples from $p(x|y)$ via standard Gaussian
process inference. We derive an MCMC procedure for the remaining
variables.
There are several methods for inference in stationary Gaussian
processes affording efficient implementation and online processing
\cite{DBLP:conf/uai/HensmanFL13,5589113,Deisenroth_ITAC_2012}. For our
purposes, we found equidistant sampling in time and matrix inversion
sufficient to provide trajectory estimates.  
We used a variant of \emph{reversible-jump MCMC}
(RJ-MCMC, \cite{green1995reversible}),  in order to infer activities. The steps we
used include Birth/Death, Split/Merge, and parameter modification for $\Delta t,t_i^s, c_i,r_i$ and the participants. The full set
of allowed steps is described in Appendix~B.
The number of burn-in iterations required depends on the number of
activities, participants, time samples, and the relative
identifiability of the activities. In the examples shown in
Section~\ref{sec:results}, $\sim 10^4$ iterations
sufficed. 

\begin{algorithm}[ht!]
\caption{Estimating activities by RJ-MCMC\label{alg_RJMCMC}}
\begin{algorithmic}[1]
\FOR{$i=1,2,\dots,N_{hist}$}
\STATE Choose a step type: birth/death, split/merge, or type, center, radius, span, start-time and participants changes, see Appendix~B. for the description of each step.
\STATE Update GP observations according to Subsection~\ref{subsec_conditioning_gp} (optional).
\STATE Compute proposed new configuration and its acceptance probability based on the model at \ref{eq:complete_prob}.
\STATE Update configuration if accepted, if $i>N_{burn-in}$, save configuration samples
\ENDFOR
\end{algorithmic}
\end{algorithm}

\vspace{-0.15cm}
\subsection{Collaborative Localization}
\vspace{-0.15cm}
Our model enables improved localization by associating data streams via
activities, even in the face of sparse GPS measurements. For a set of
trajectories $x$ we have:
\begin{align}
\label{eq:localization}
 p(x|A,y,\theta) = p(A,x|y,\theta)/p(A|y,\theta) \propto p(A|x)p(x|y,\theta)
\end{align}
via Bayes rule and the dependency structure. 
%
%
%
Marginalizing over
the sampled set
of configurations, we can reduce uncertainty based on interactions with other actors, as demonstrated in Section \ref{subsec:res_localization}.
%
%

%
%
\vspace{-0.15cm}
\subsection{Collaborative Summarization}
\vspace{-0.15cm}
\label{subsec_inf_summarization}
%
An important aspect of collaborative summarization that we demonstrate is keyframe selection that emphasizes activities of interest.
We do so using a farthest-point-sampling approach (FPS, \cite{Gonzalez85,Hochbaum85}), with an appropriate
metric between frame pairs (see Appendix~C for full details). 
We say that two frames disagree with respect to an activity
if one of them was sampled as a part of this activity, and the other was not.

We denote by $d_{AC}(f_i,f_j)$ a measure of disagreement between frames $f_i,f_j$ marginalized over activity configurations.  
The distance between two frames is a positive linear combination of $d_{AC}$ as well as the feature-space, temporal, and participant identities distances. It is a metric, as required for FPS.
%
%
Sampled keyframes are reordered according to time. When selecting keyframes, we
pick only from those that are likely to have participated in the relevant
activities or in the relevant actors' datastream.
The selected frames can be used for a video summary, or a map-based summary using their inferred locations.

 \begin{algorithm}[ht!]
\caption{Keyframe selection for collaborative activities\label{alg_summary}}
\begin{algorithmic}[1]
\STATE Initialize counts to be $0$
\FOR{All configurations $\{A_i\}$ and activities $i$}
\FOR{All actors $a$ and frames $t$ }
\STATE If actor's frame is within activity $i$'s span, add a vote.
\ENDFOR
\ENDFOR
\STATE Remove keyframes with low votes count, compute frame-pair distances.
\STATE Sample keyframes via FPS, re-order keyframes according to time.
\end{algorithmic}
\end{algorithm}
\textbf{Video Summarization} For the case of videos, tradeoff between information variability and film consistency is used, similar to  Arev \etal \cite{Arev:2014:AEF:2601097.2601198} (but using only epipolar geometry and average image motion in the absence of 3D metric reconstruction). 

\textbf{Map-based Summarization}
For a summary image, we overlay relevant frames on a map of the area, by placing
keyframes inside the area defined by each activity's center and radius. We use FPS in order to place the frames uniformly.  




\section{Results}
\label{sec:results}
We first explore the use of our algorithm on a synthetic dataset to ensure
accurate detection of activities under model circumstances, then proceed to
demonstrate activity inference, summarization, and localization in the case of
real-world wearable data.
 \vspace{-0.2cm}
\subsection{Activity Inference}
\begin{figure}[h!]
\centering
\begin{minipage}{0.4\linewidth}
\centering
\includegraphics[width=\linewidth]{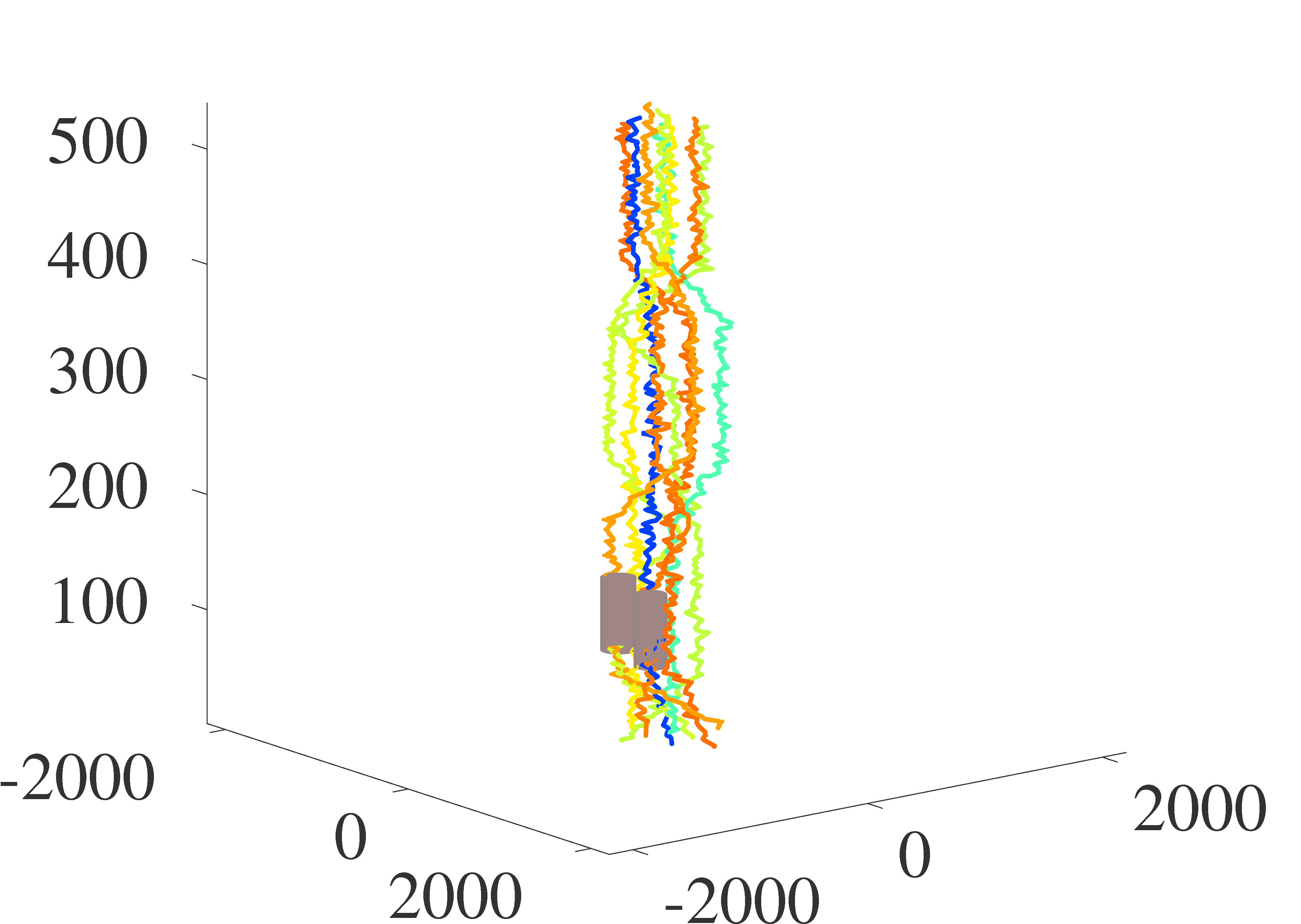} 

(a)
\end{minipage}
\hfill
\begin{minipage}{0.4\linewidth}
\centering
\includegraphics[width=\linewidth]{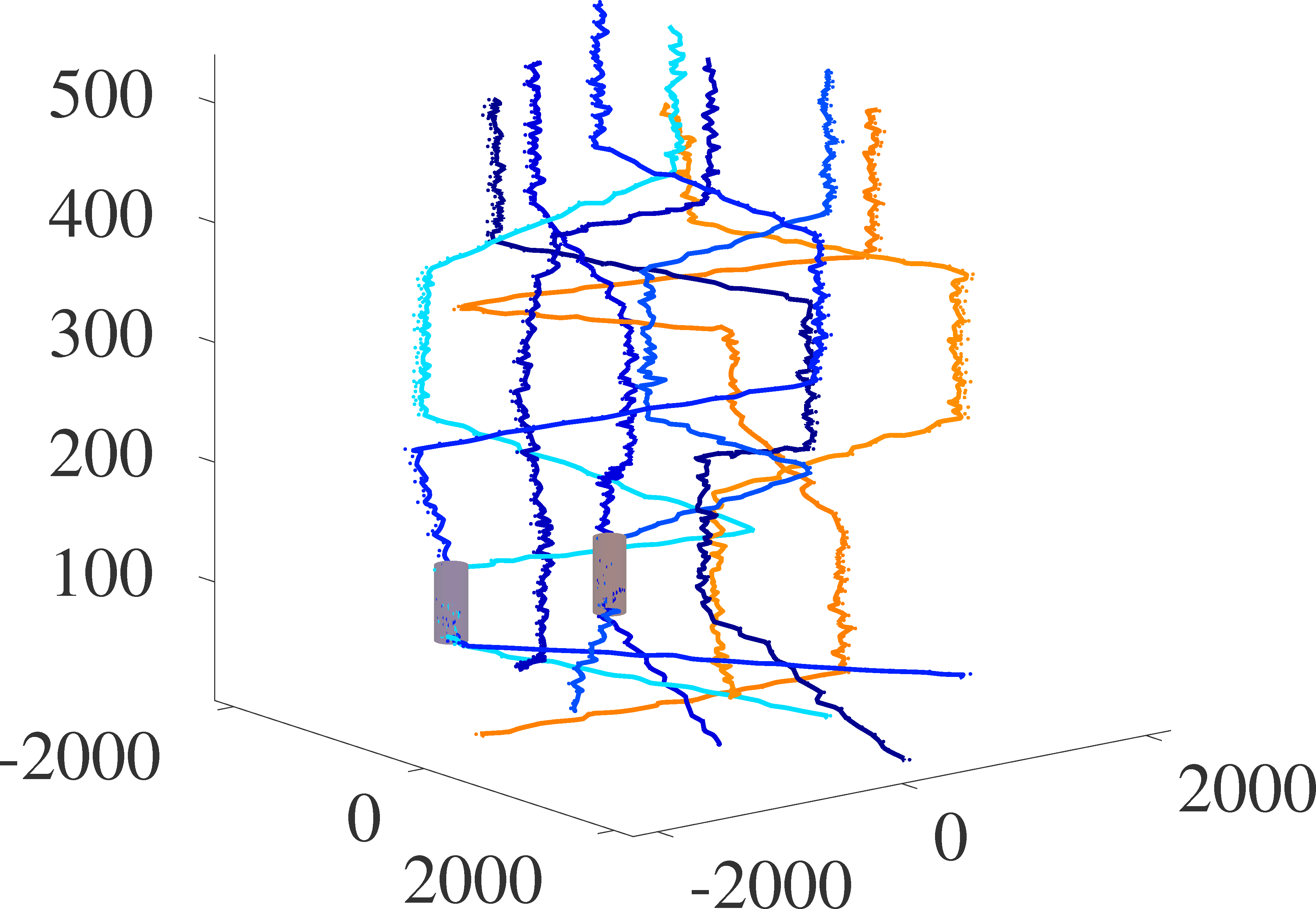} 

(b)
\end{minipage}

\vspace{1cm}
\begin{minipage}{0.8\linewidth}
\centering
\includegraphics[width=\linewidth]{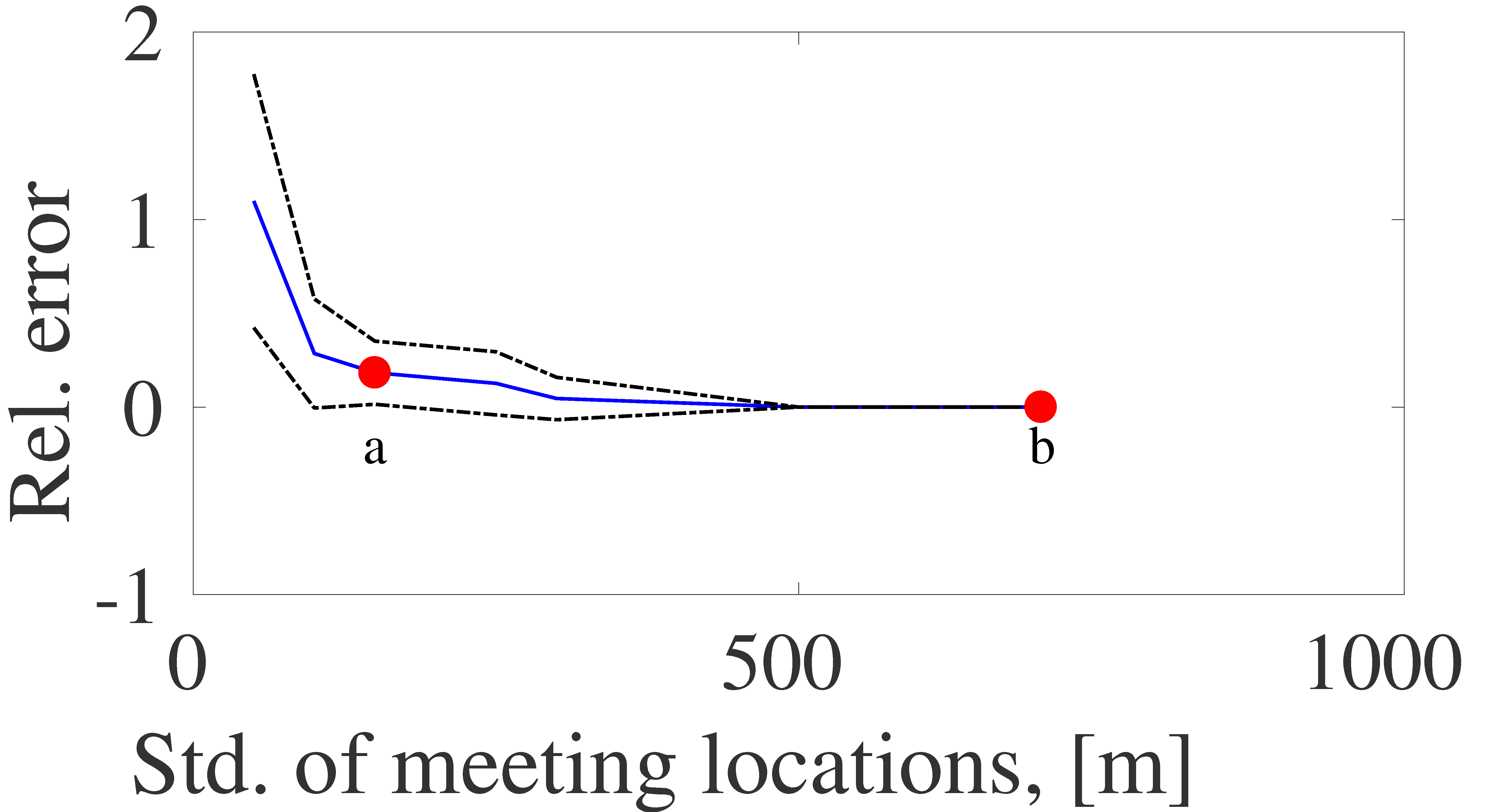} 

(c)
\end{minipage}

 \caption{\label{fig:synthetic} \figurefontsize Example results from synthetic experiment of
   meetings among eight actors. (a),(b) Left subfigures: example trajectories and detected configurations, for different distributions of meetings' locations. The $x,y$
   axes represent spatial dimensions and the vertical axis represents time
   shown from two viewpoints. The std. of activity locations in (a),(b) is 150 and 700 meters, respectively, compared to GPS noise and activity mean radius of 30 meters.  
   Different colored ensembles of curves represent five
   estimated tracks from each participant, based on GPS observations (marked as dots). Cylinders mark the
   spatio-temporal span of detected activities.  Right: the relative error in
   the estimation of the number of activities, averaged over 20 trials, black dashed lines mark $\pm 1$ std. curves. The 2 red dots show the std. for Subfigures a,b. As can be seen, algorithm manages to detect the activities even when they are barely discernible in the GPS observations. }
 \end{figure}

 \vspace{-0.2cm}
\label{subsec:res_activity}
We now demonstrate the result of inferring plausible configurations of
activities on a synthetic dataset generated according to the following
structure: at each turn, each of eight actors can either go to one of several
meeting places, or go to their own random location. If two or more actors are
at the same location, this constitutes an activity. We test the inference
procedure on this model. When meeting places are well-separated from each other
(compared to the observation noise and typical meeting radius), we can infer
activities without difficulties.  Figure~\ref{fig:synthetic}
explores the increase in detection errors as a function of the standard
deviation of the meeting locations, over multiple realizations. The distribution of meeting locations should be contrasted with
a measurement noise of 30 meters, and meeting radius distributed with $\mu_r
= 30$, and a meeting-times prior set to log-normal with parameters
$\mu_t=60,\sigma_t=0.05$.  Looking at detected activities, the main source
for false detection occurs when actors unintentionally cross paths for prolonged periods of time. 

We now demonstrate activity inference on a real dataset, available at
\cite{VideoDatasetReviewVersion}.  This dataset includes four actors walking for
half an hour in an urban area spanning 600 meters, and meeting each
other, either on the street (activity type 1) or inside a coffeeshop (activity type 2). 
During a training phase, we construct our feature priors based on a subset of images from such locations so as to identify the two activity types with this semantic categorization.  
The actors are wearing
GoPro cameras that provide a partial, egocentric video, with over $2300$ classified face tracklets in the videos.
The error rate for face detection in the video was
$14\%$, giving us a noisy, but informative, signal. 

\begin{figure}

\centering

\begin{minipage}{0.47\linewidth}
  \begin{minipage}{0.44\linewidth}
  \includegraphics[width=1\linewidth]{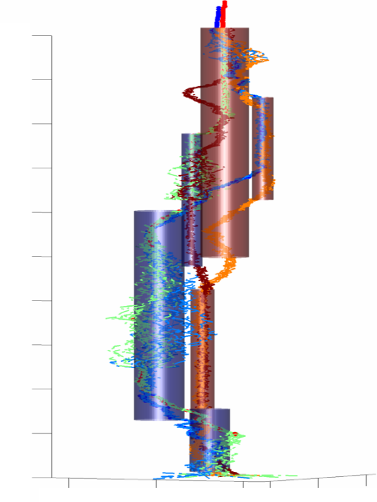} 
  \end{minipage}
  \hfill
  \begin{minipage}{0.45\linewidth}
  \includegraphics[width=\linewidth]{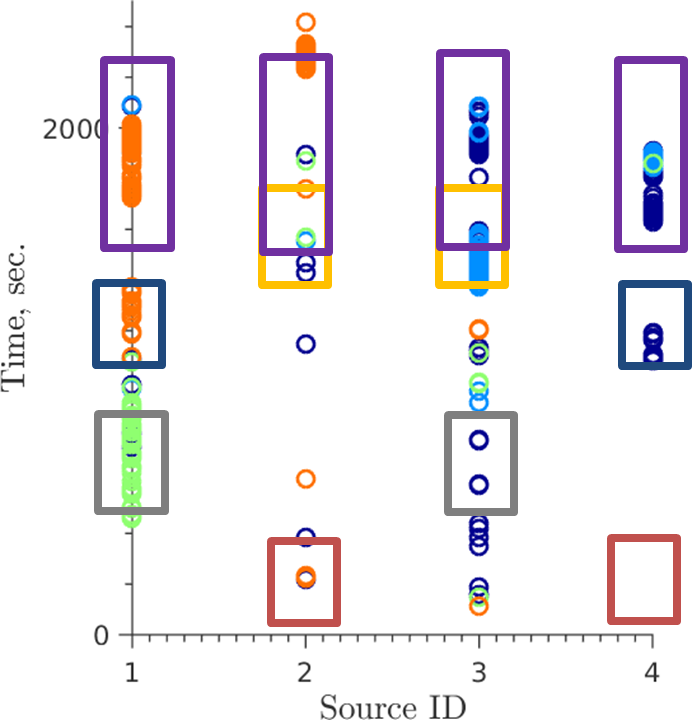} 

  \hfill
  \begin{minipage}{0.19\linewidth}
  \includegraphics[width=\linewidth]{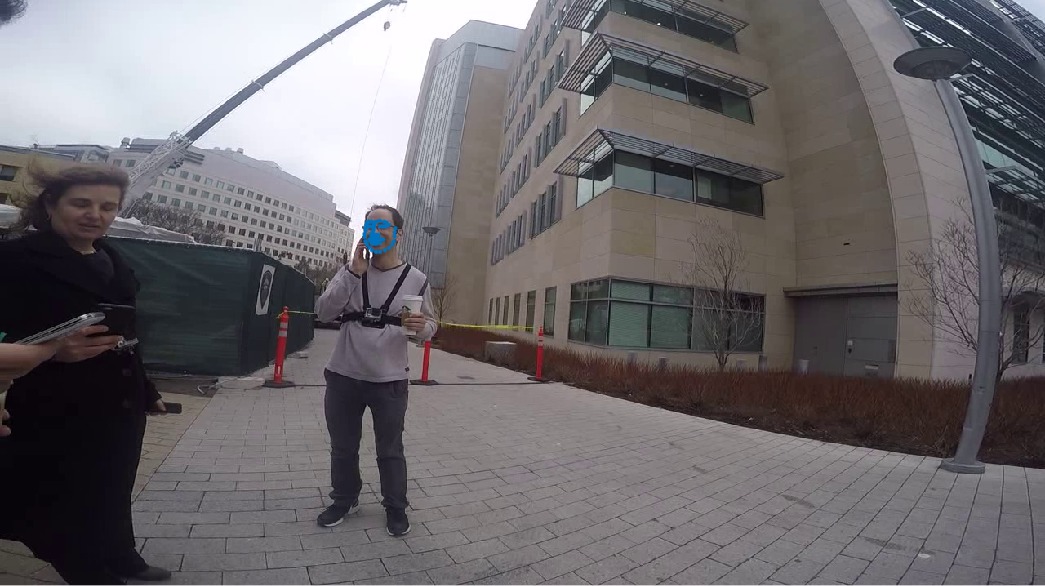} 
  \end{minipage}
  \hfill
  \begin{minipage}{0.19\linewidth}
  \includegraphics[width=\linewidth]{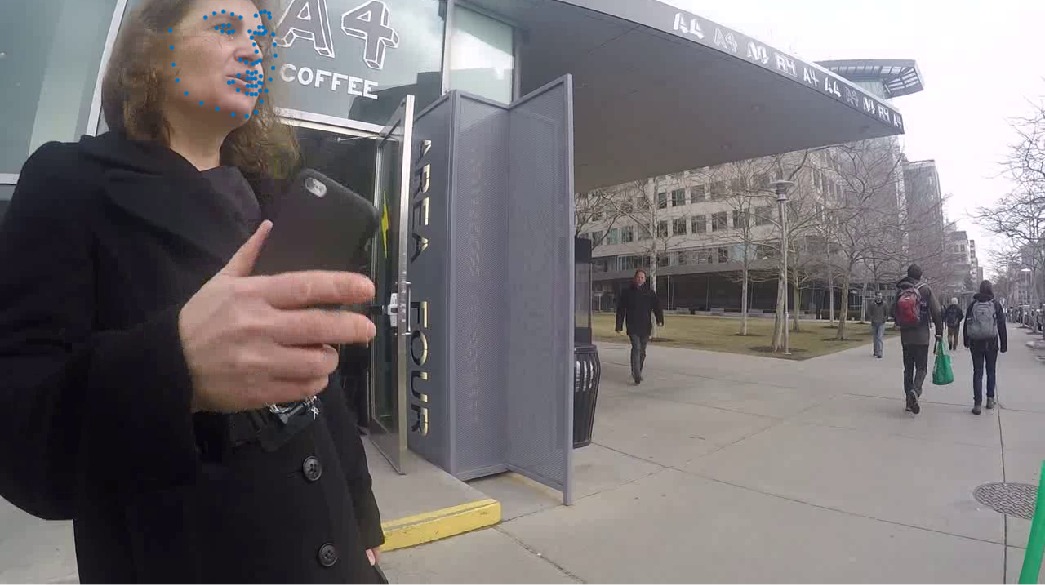} 
  \end{minipage}
  \hfill
  \begin{minipage}{0.19\linewidth}
  \includegraphics[width=\linewidth]{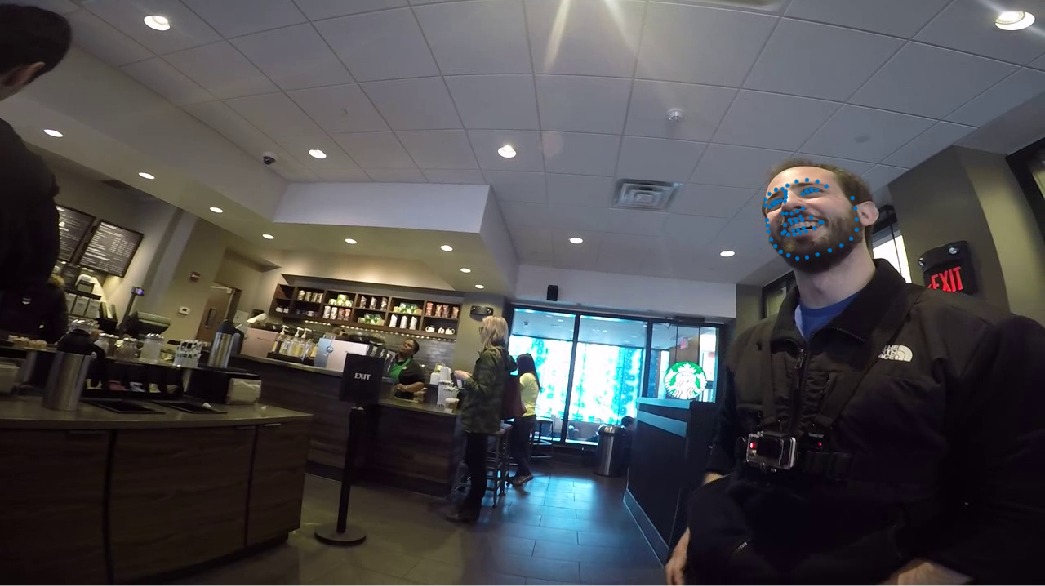} 
  \end{minipage}
  \hfill
  \begin{minipage}{0.19\linewidth}
  \includegraphics[width=\linewidth]{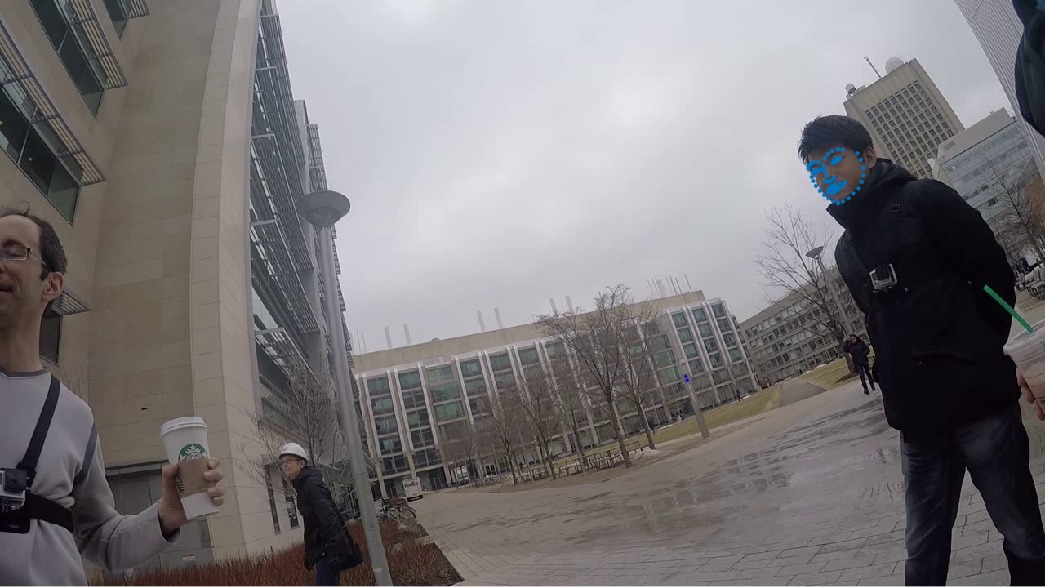} 
  \end{minipage}
\end{minipage}

\centering

\vspace{0.5cm}
(a)

\end{minipage}
\hfill
\begin{minipage}{0.45\linewidth}
\centering

\begin{minipage}{1\linewidth}
\centering
\includegraphics[width=\linewidth]{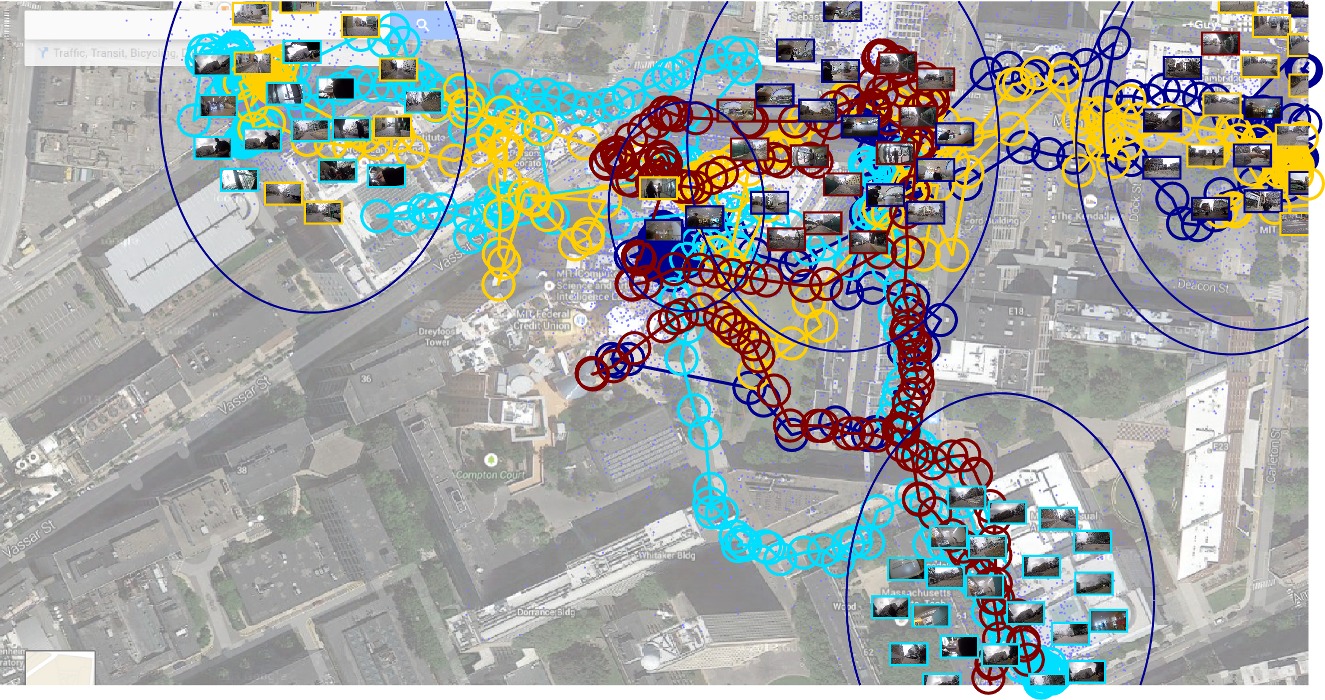} 
\end{minipage}

\begin{minipage}{0.23\linewidth}
\includegraphics[width=\linewidth]{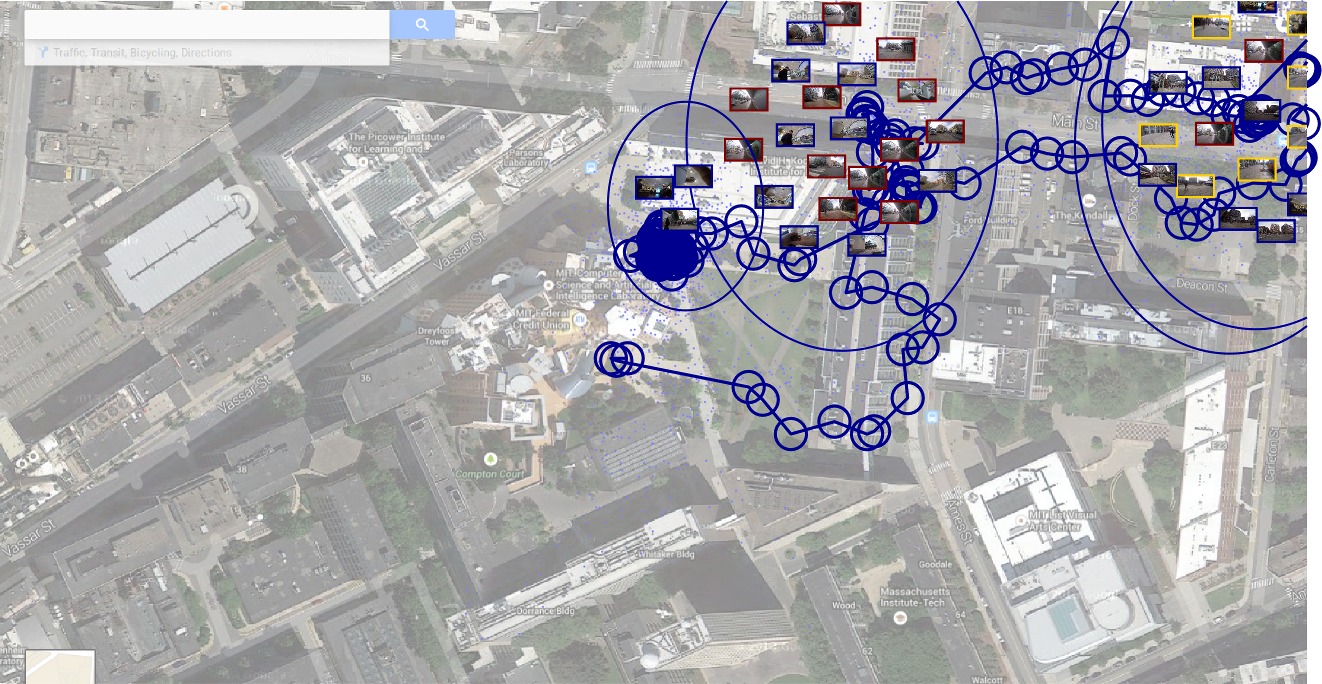} 
\end{minipage}
\begin{minipage}{0.23\linewidth}
\includegraphics[width=\linewidth]{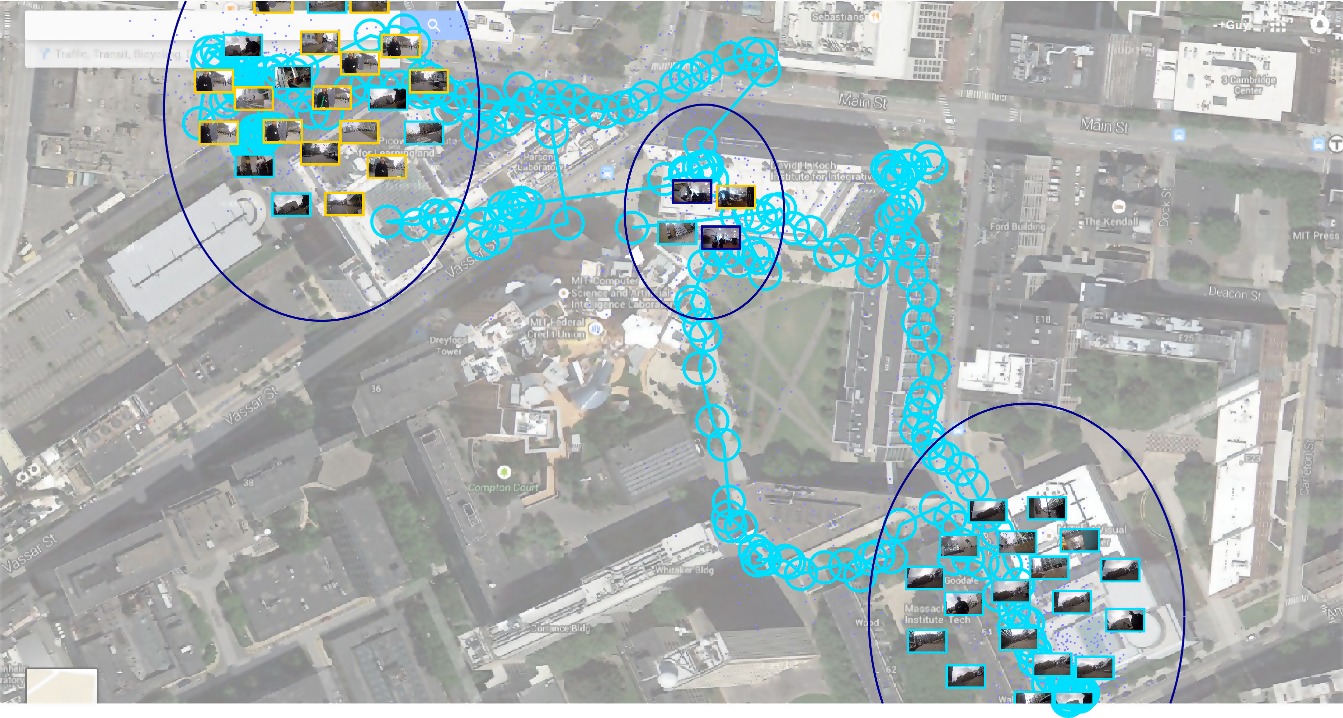} 
\end{minipage}
\begin{minipage}{0.23\linewidth}
\includegraphics[width=\linewidth]{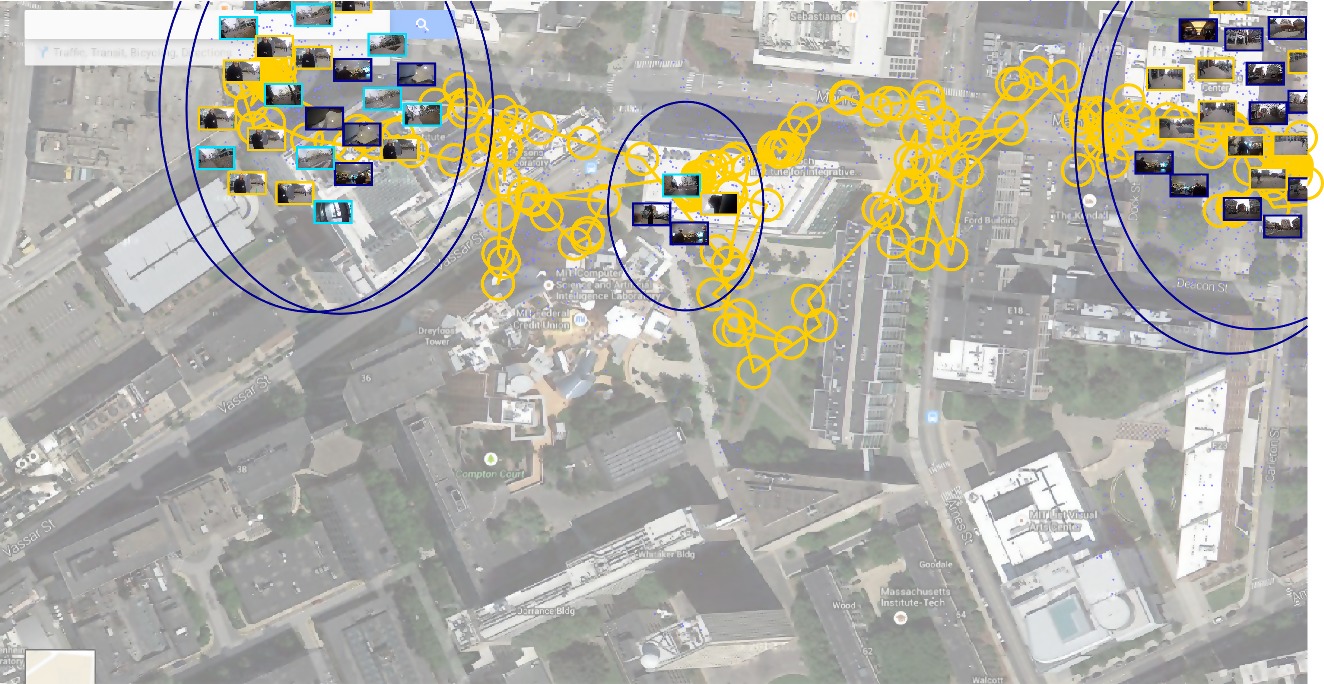} 
\end{minipage}
\begin{minipage}{0.23\linewidth}
\includegraphics[width=\linewidth]{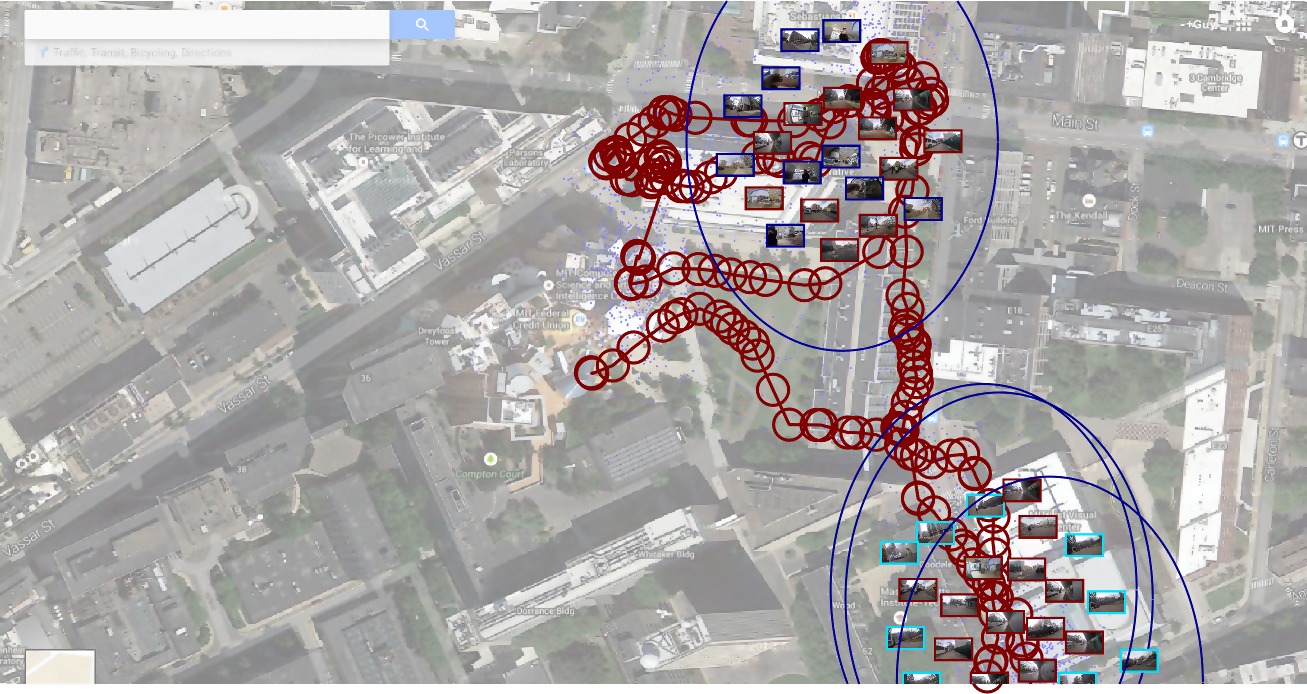} 
\end{minipage}

\centering

\vspace{0.2cm}
(b)

\end{minipage}

\caption{\figurefontsize\label{fig:activity_detection} a) Left: Activities detected on our dataset, with four actors walking and interacting for approx. 30 minutes. Blue cylinders represent indoor meetings, whereas red cylinders represent outdoor meetings.
 Right: a time plot of detected faces in the different actors' video feed, overlayed with a ground-truth of the meetings that took place. As can be seen, all meetings were detected, and are to some extent evident in the face detection streams.
\label{fig:summary_map} b) Top: summarization of actors activities with respect to the map, based on the GPS and video footage of $4$ actors. 
 Colored trajectories represent actors GPS location measurements, using the same actor color as Figure~\ref{fig:summary_storyline}. Circles mark the trace of detected activities, with representative frames displayed on top. 
 Bottom: the individual four actors storylines, with activities depicted based on the pooled actors' video footage. The map and images have been blurred for review.}

\end{figure}

During some of the meetings, GPS reception was poor with gaps ranging from one
to nine minutes in duration, in the GPS coordinate measurements of some
participants.  Furthermore, ten minutes of video were removed from participant
number four, in order to test data pooling from different sources.  As can be shown
in Figure~\ref{fig:activity_detection}, all meetings are detected in a stable
manner, despite missing GPS. This dataset is simple enough to be easily understood, and yet captures a lot of the difficulties in understanding and summarizing collaborative interactions in a realistic urban environment.
An additional example, also included in \cite{VideoDatasetReviewVersion}, with $11$ participants is shown in Subsection~\ref{sec_larger_example}. While the urban environment and multiple activities makes interpretation of results more difficult in this example, we are still able to detect most of the activities
and demonstrate how the approach scales up for more complex data.
  
\subsection{Summarization}
\label{subsec:res_summarization}
\begin{figure*}
\centering
\begin{minipage}{1\linewidth}
\includegraphics[width=\linewidth]{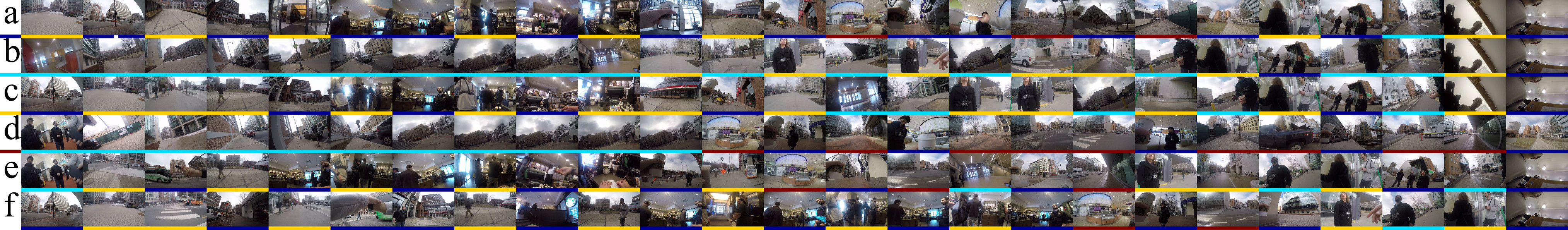} 
\end{minipage}

 \caption{\figurefontsize\label{fig:summary_storyline} Rows (a)-(d): summarization for actors 1-4, utilizing footage from likely participants in each actor's activities, going left to right. See left size colors legend and frame color to mark the frame source. 
 Row (e): A combined storyline  of the four actors, using images from all of them. Row (f) All four actors, but with emphasis on activities that are likely a ``coffeeshop''. 
 Note how in the beginning of the timeline, (d) takes footage from participant (b) during their activity, whereas actors (a),(c) share footage. 
 The ending is similar for most participants due to a final joint meeting. 
 }
\end{figure*}

In Figure~\ref{fig:summary_storyline} we demonstrate the summarization of actors' data from all four actors. Rows (a)-(d) summarize individual actors' timeline, pooling footage from other actors who participated in that actor's activities.
The fifth row summarizes the data from all four actors and activities, whereas the sixth row summarizes all actors but favoring frames from ``coffeeshop'' activities.
Summarization is done while attempting to balance content and time variability, and activity representation, as described in Subsection~\ref{subsec_inf_summarization}. We highlight the beginning of actor 4's storyline. This actor's video data for the first few minutes was removed from the set, and yet, the algorithm has completed it due to the detection of a meeting between actors 3,4.

In Figure~\ref{fig:summary_map} we demonstrate the summarization of activities on a map. The GPS measurement locations are shown for each actor on the map, and demonstrate the noisy raw data used to estimate the locations and infer the activities. Green circles represent a set of detected activities.
Inside each circle, several images are placed and represent main parts of the meeting.


 \vspace{-0.2cm}
\subsection{Localization}
\begin{figure}[h!]
\centering
 \begin{minipage}{0.06\linewidth}
 \ 
 \end{minipage}
 \begin{minipage}{\gpsFigureWidth\linewidth}
 \centering
 \includegraphics[width=\linewidth]{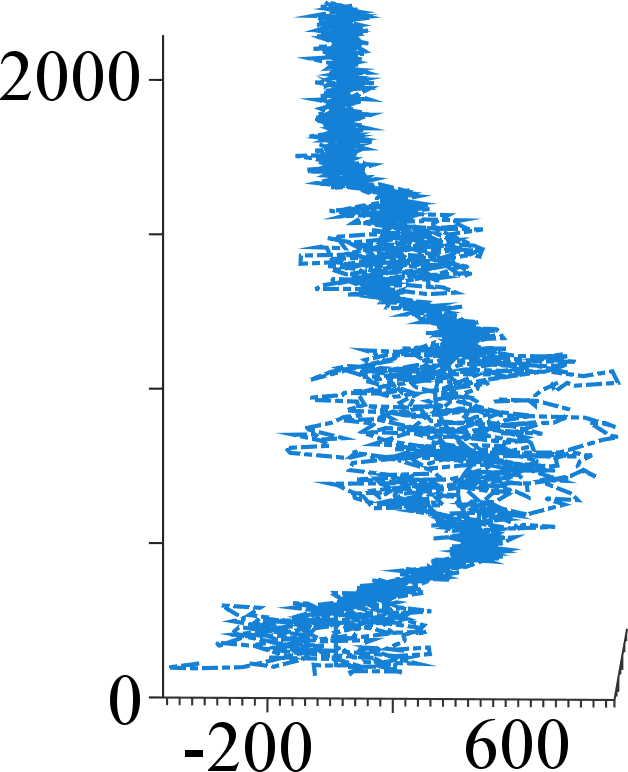} 
 
 (a)
 \end{minipage}
 \begin{minipage}{\gpsFigureWidth\linewidth}
 \centering
 \includegraphics[width=0.86\linewidth]{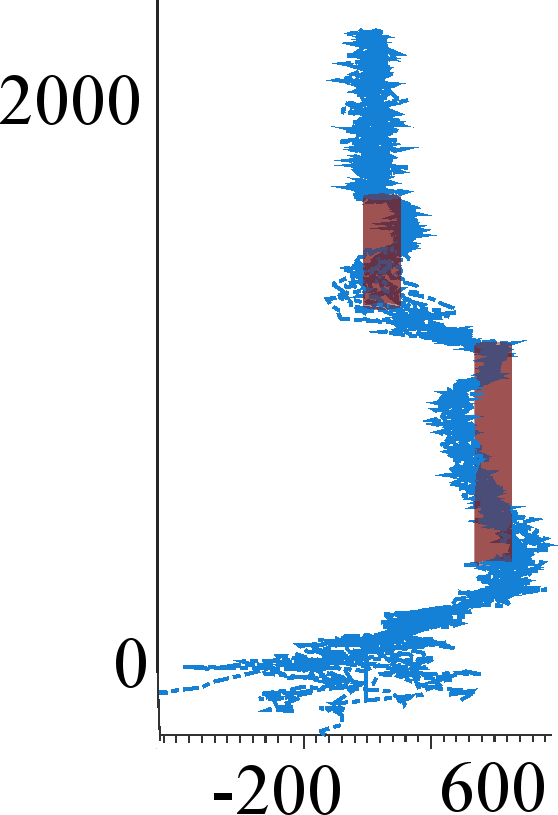} 
 
 (b)
 \end{minipage}
 \hfill
 \begin{minipage}{\gpsFigureWidth\linewidth}
\centering
\includegraphics[width=0.62\linewidth]{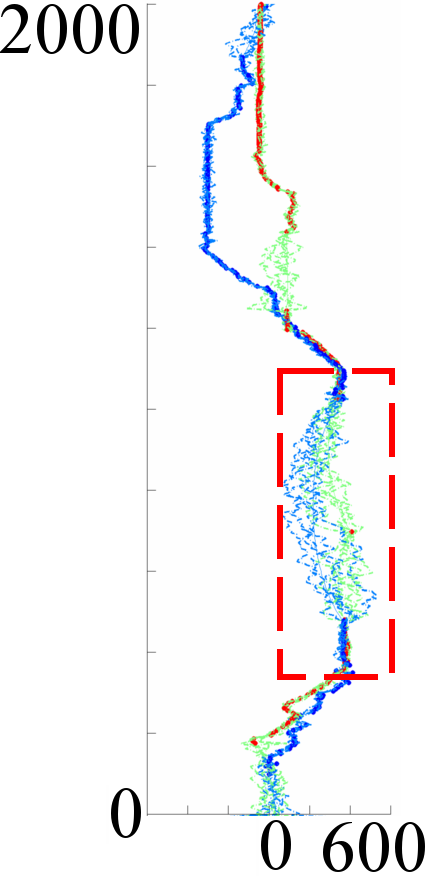} 

 (c)

\end{minipage}
\begin{minipage}{\gpsFigureWidth\linewidth}
\centering
\includegraphics[width=0.52\linewidth]{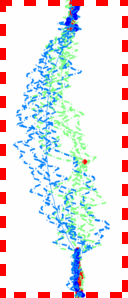} 

 (d)

\end{minipage}
 \begin{minipage}{0.05\linewidth}
 \ 
 \end{minipage}
\hfill


\centering
 \begin{minipage}{0.06\linewidth}
 \  
 \end{minipage}
 \begin{minipage}{\gpsFigureWidth\linewidth}
 \centering
 \includegraphics[width=0.8\linewidth]{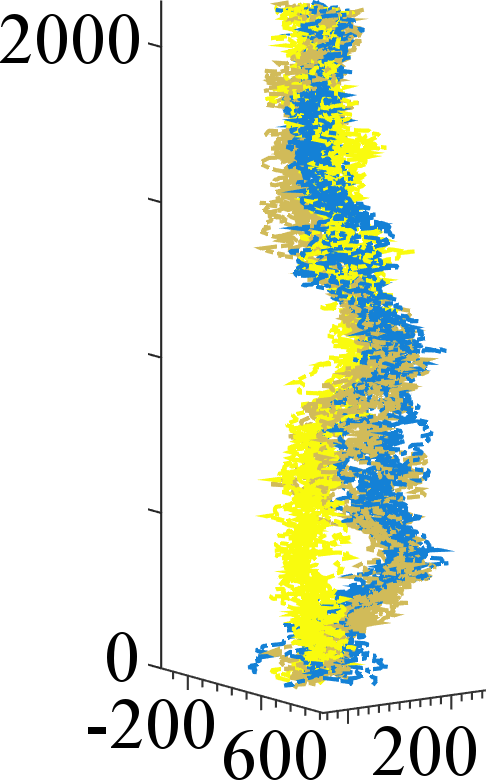} 
 
  (e)
 \end{minipage}
 \begin{minipage}{\gpsFigureWidth\linewidth}
 \centering
 \includegraphics[width=0.8\linewidth]{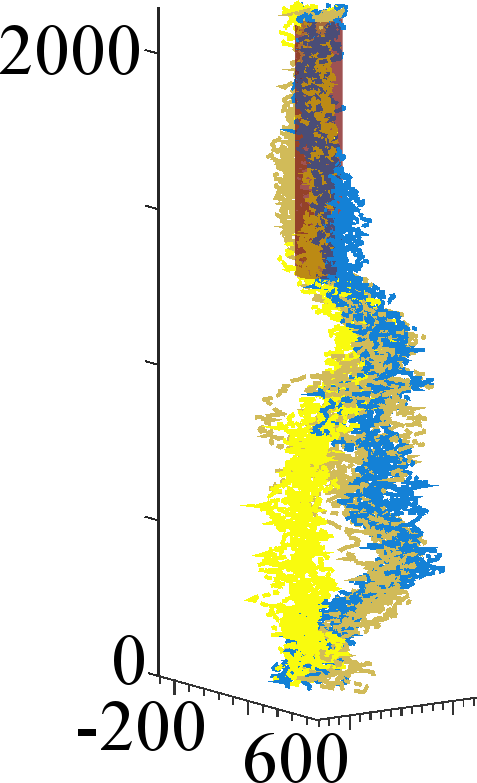} 
 
  (f)

 \end{minipage}
 \hfill
%
\begin{minipage}{\gpsFigureWidth\linewidth}
\centering
\includegraphics[width=0.62\linewidth]{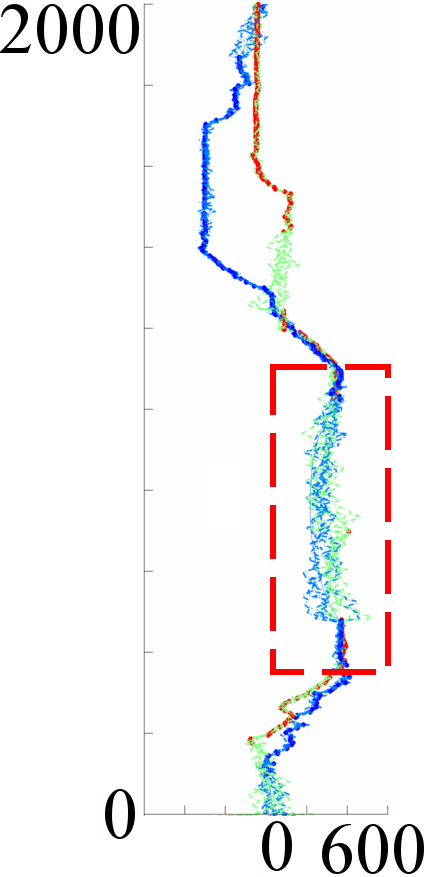} 

 (g)

\end{minipage}
\begin{minipage}{\gpsFigureWidth\linewidth}
\centering
\includegraphics[width=0.52\linewidth]{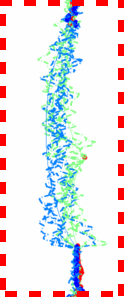} 

 (h)

\end{minipage}
 \begin{minipage}{0.05\linewidth}
 \ 
 \end{minipage}
\hfill


 \caption{\figurefontsize\label{fig:localization} (a),(b): Estimated GP trajectories for a single actor, 
 as $(x_1,x_2,t)$, with units $([\mbox{m}],[\mbox{s}])$, before and after conditioning on sampled activities (shown as a maroon cylinders). 
   Note the
   divergence of samples once GP measurements are
   missing, where actors are inside a building.  
   (e),(f): The trajectories of two participants before and after conditioning on an activity.
(c),(d) and (g),(h)
Max-likelihood trajectories of two participants holding a meeting (around 500 seconds, on the $z$ axis) -- 
 full and zoomed view of $x$ conditioned on observations $y$, followed by $x$ conditioned on the detected activities configuration $\{A_i\}$ and $y$. 
 Clearly by conditioning on $\{A_i\}$ we can better localize the two people despite GPS-denial. 
 }
 \vspace{-0.3cm}
\end{figure}

 \vspace{-0.2cm}
\label{subsec:res_localization}
Our model allows us to better localize actors from partial GPS measurements, through the GP samples, and the posterior likelihood of the trajectories.
We use Equation~\ref{eq:localization} to visualize a sampled trajectory set for each actor.
While ignoring trajectories dependence, this visualizes likely trajectories for each actor, as shown in Figure~\ref{fig:localization} for two actors
in a GPS-denied cafe for about ten minutes, starting at $t=500$. While their location is less certain, 
conditioning on the sampled configurations allows us to better localize them better. The average standard deviation of the two participants $(1,3)$ during that meeting dropped from $(73,90)[m]$ to $(53,57)[m]$, respectively.
The same phenomenon can be seen for shorter GPS denial segments as well, around $t=1250$, with reduction in standard deviation for participants $(1,4)$ from $(66,68)[m]$ to $(53,44)[m]$, respectively. This shows halving of the variance in such cases.

While for this example linear interpolation may work in a low-noise setting, our GP model allows activities where participants are moving together (\eg field-trip or running group).

 \vspace{-0.2cm}
\subsection{Prior for Face Recognition}
 \vspace{-0.2cm}
\label{subsec:res_face_recognition}
We compute a conditional probability of facial features given participant identity and the maximum-likelihood configuration, as described in Subsection~\ref{subsec:infinite_model}. 
This lowers the face recognition error on average by $5.5\%$, demonstrating the utility of a unified multimodal model for explaining the data. 

\begin{figure}[htbp]

\centering
\centering
\begin{minipage}{0.45\linewidth}\centering
\includegraphics[width=\linewidth]{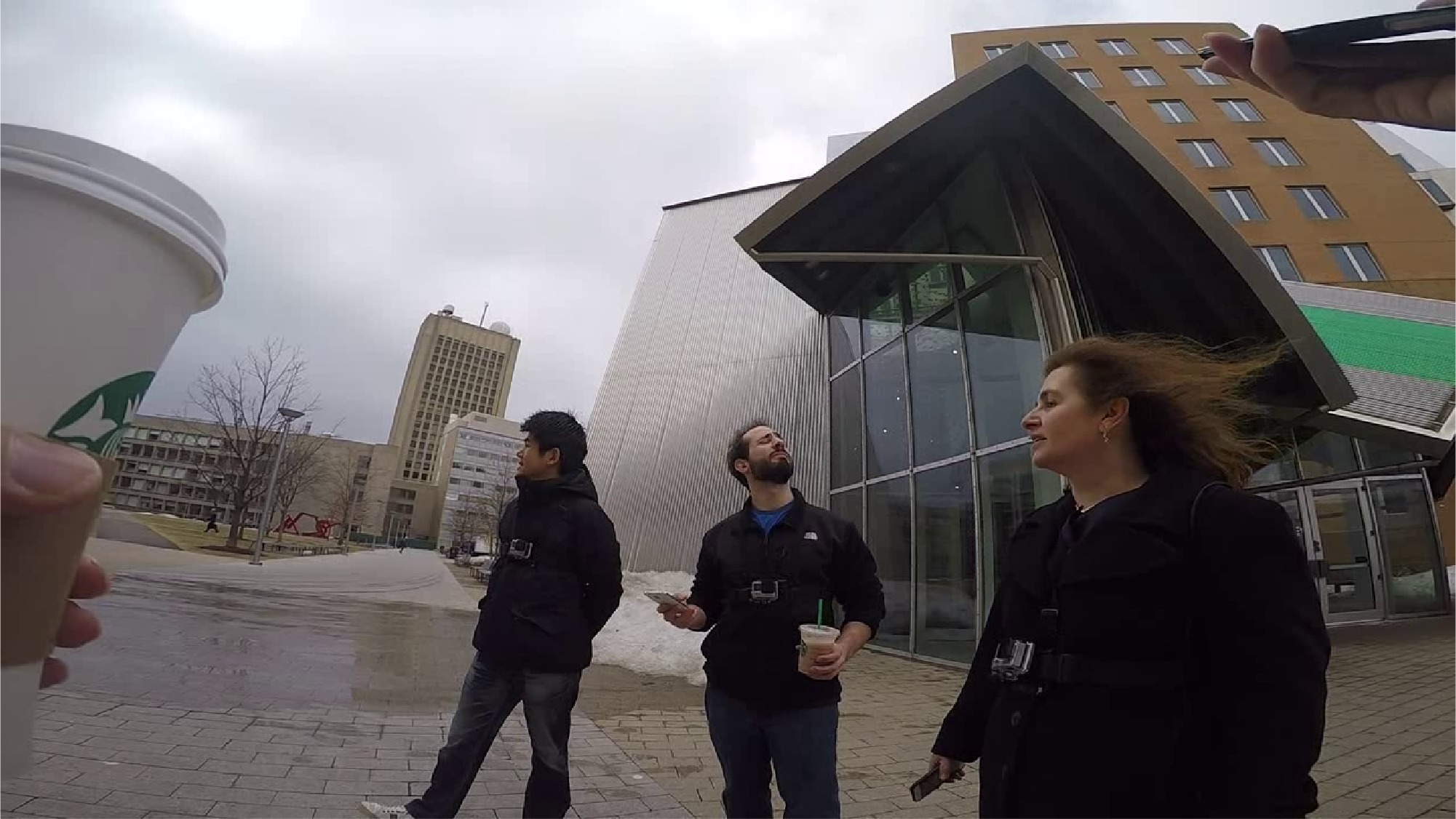}

\footnotesize (a) Classifier: unrecognized. Posterior: Participant 2.
\end{minipage}
\begin{minipage}{0.45\linewidth}\centering
\includegraphics[width=\linewidth]{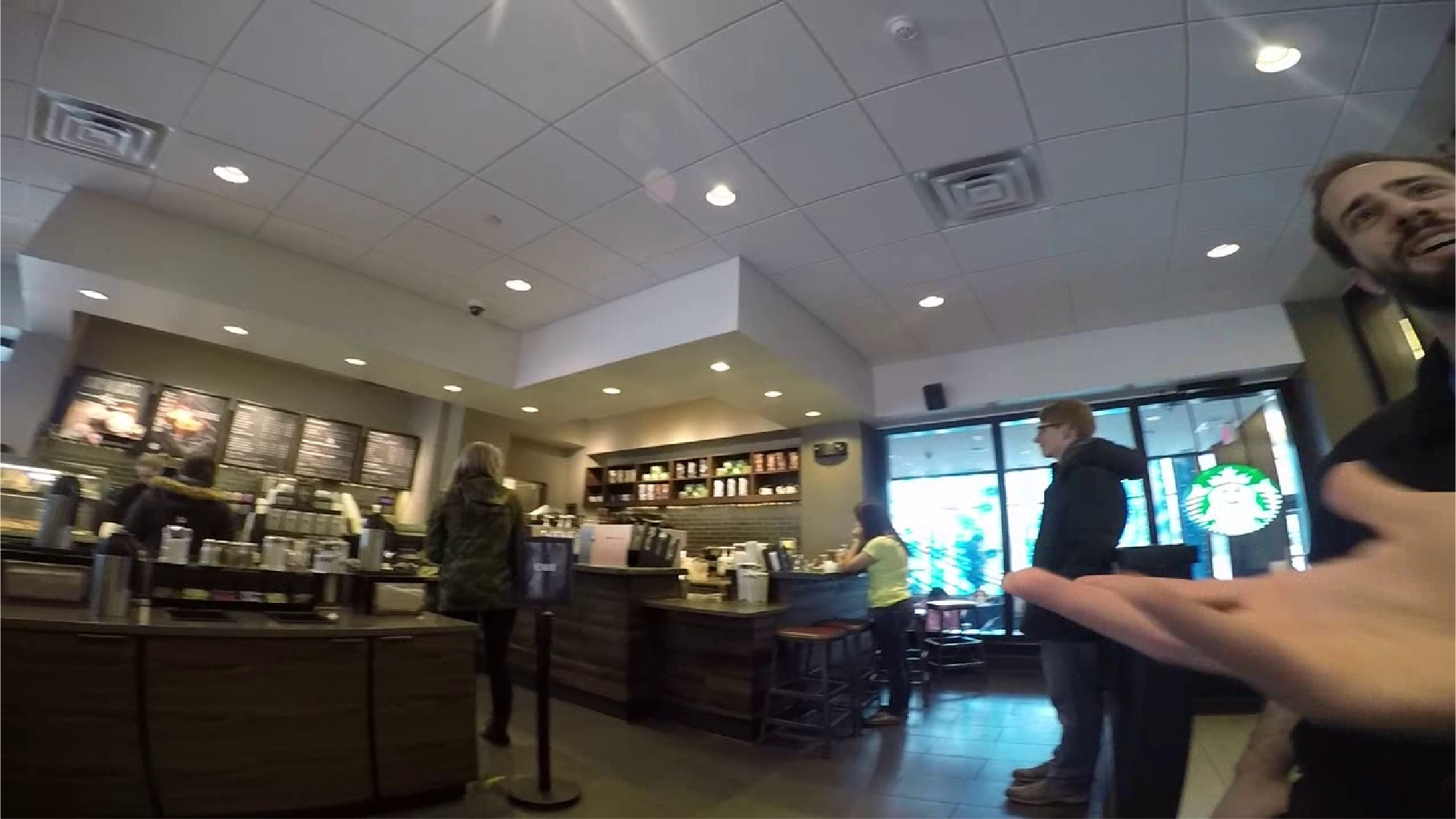}

\footnotesize (b) Classifier: participant 1. Posterior: participant 3.
\end{minipage}
\begin{minipage}{0.45\linewidth}\centering
\includegraphics[width=\linewidth]{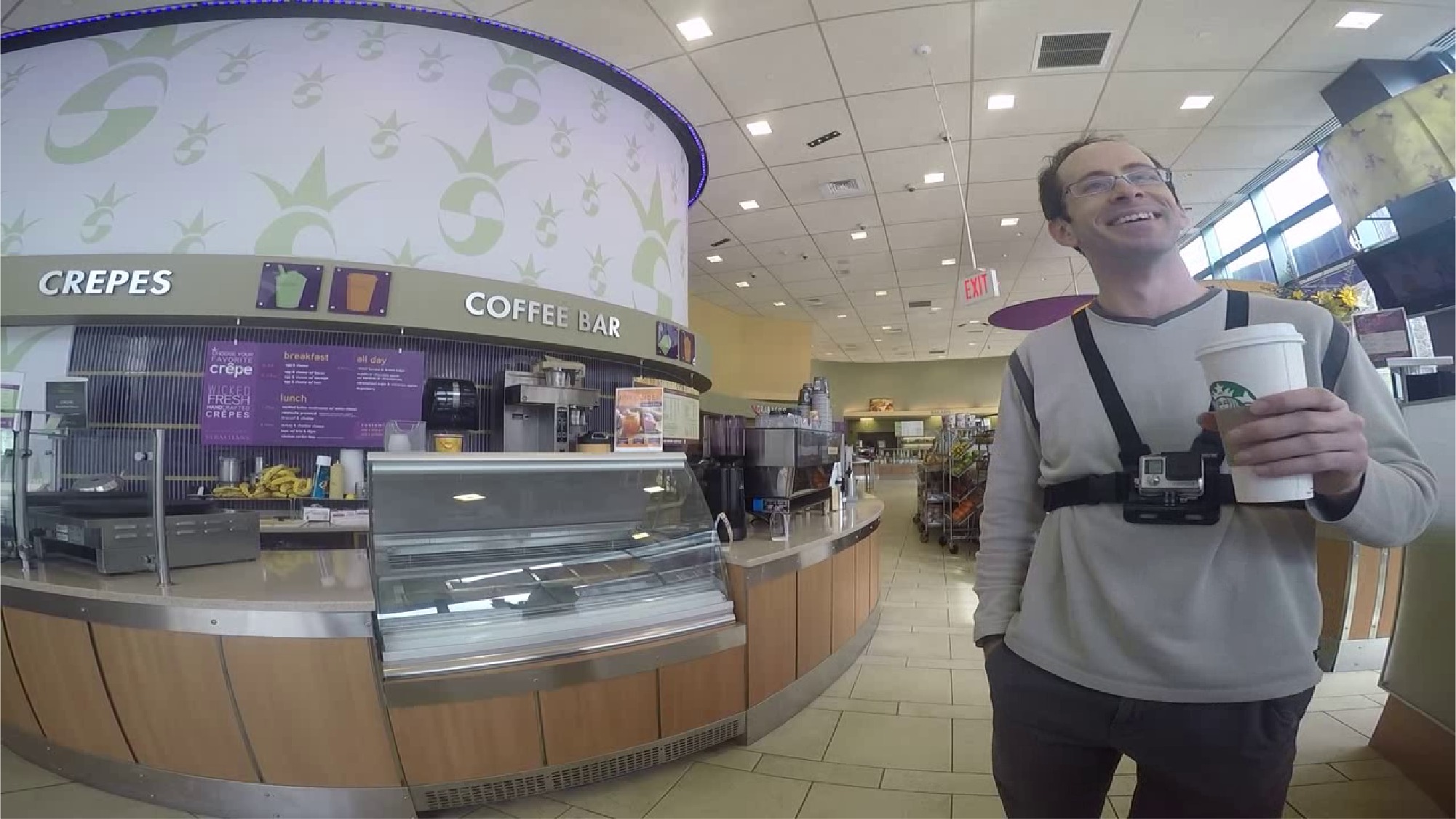}

\footnotesize (c) Classifier: participant 3. Posterior: participant 1.
\end{minipage}
\begin{minipage}{0.45\linewidth}\centering
\includegraphics[width=\linewidth]{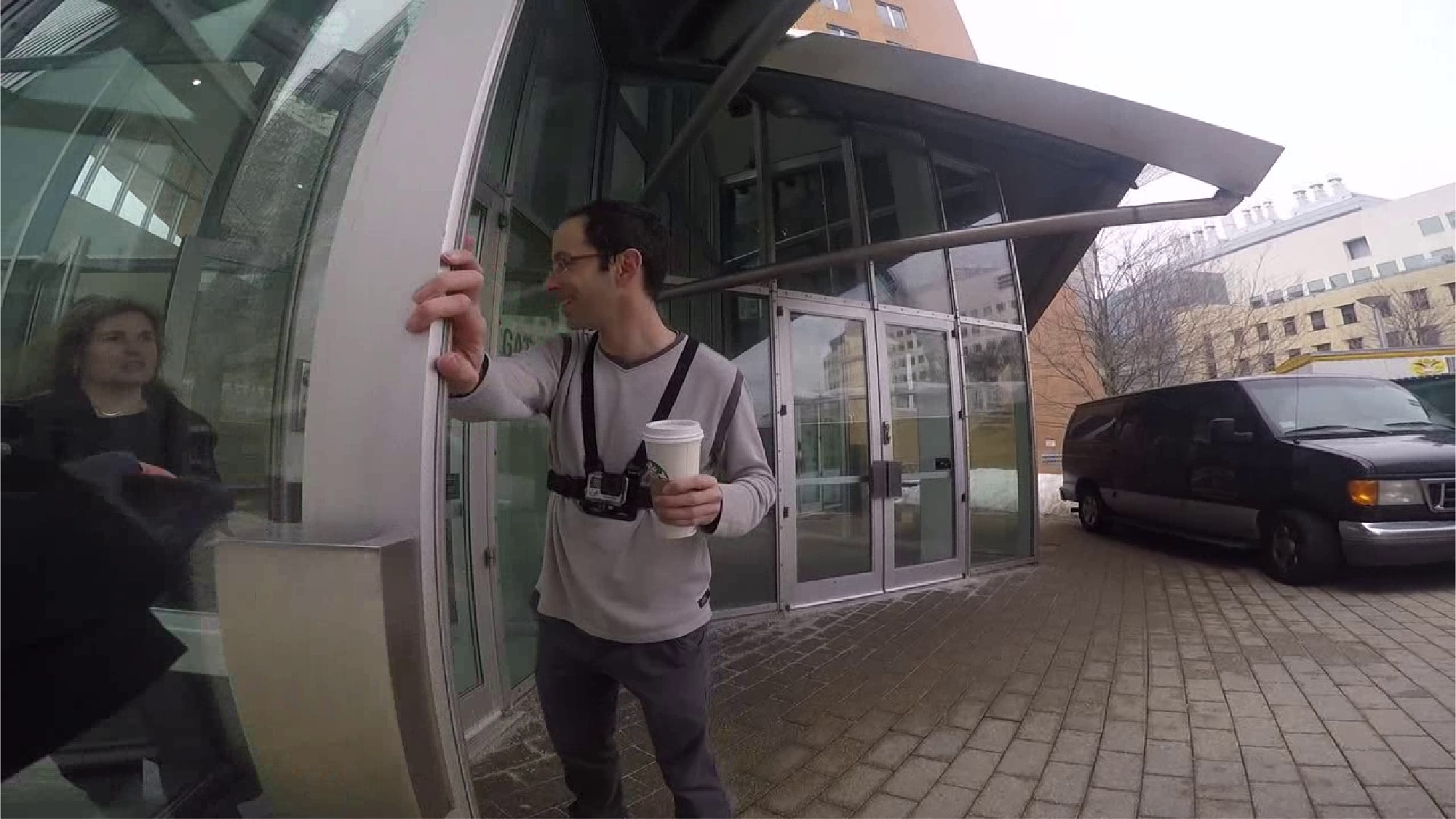}

\footnotesize (d) Classifier: unrecognized. Posterior: Participant 2.
\end{minipage}

\caption{Example face correction, based on the detected activities. Some examples were inherently difficult to recognize despite rectification and alignment, due to partiality (a,b) and reflections (d).  \label{fig:face_correction}}
\end{figure}

In Figure~\ref{fig:face_correction} we demonstrate several examples where incorrect face recognitions were corrected, conditioned on the inferred activities. 
While there is no guarantee the conditional probability always improves recognitions, in our specific case we did not see such examples. 
While the face recognition network used is relatively simple, the factors leading to incorrect labelling in those cases are relevant 
for even state-of-the-art classifiers \cite{DBLP:conf/cvpr/TaigmanYRW14}, and include partial occlusions and reflections as a face is seen through a glass door.

Assuming independent activities $A_i$, we write the probability of a specific detection's identity to be
\begin{align}
\label{eq_multiple_activities_identity}
 p(a|\{A_i\}) \propto \prod_i p(a|A_i),
\end{align}
where $p(a|A_i)$ is uniform for all activity participants, with a small non-zero value for non-participants. Plugging Equation~\ref{eq_multiple_activities_identity} into Equation~11 in the main paper results in an additive term for the log-likelihood. 
Normalizing w.r.t the partition function gives us the modified face detection probabilities.

\subsection{Larger-scale Scenario}
\label{sec_larger_example}
\begin{figure}[htbp!]
\vspace{0.5cm}
\centering
\begin{minipage}{0.45\linewidth}
\includegraphics[width=1\linewidth]{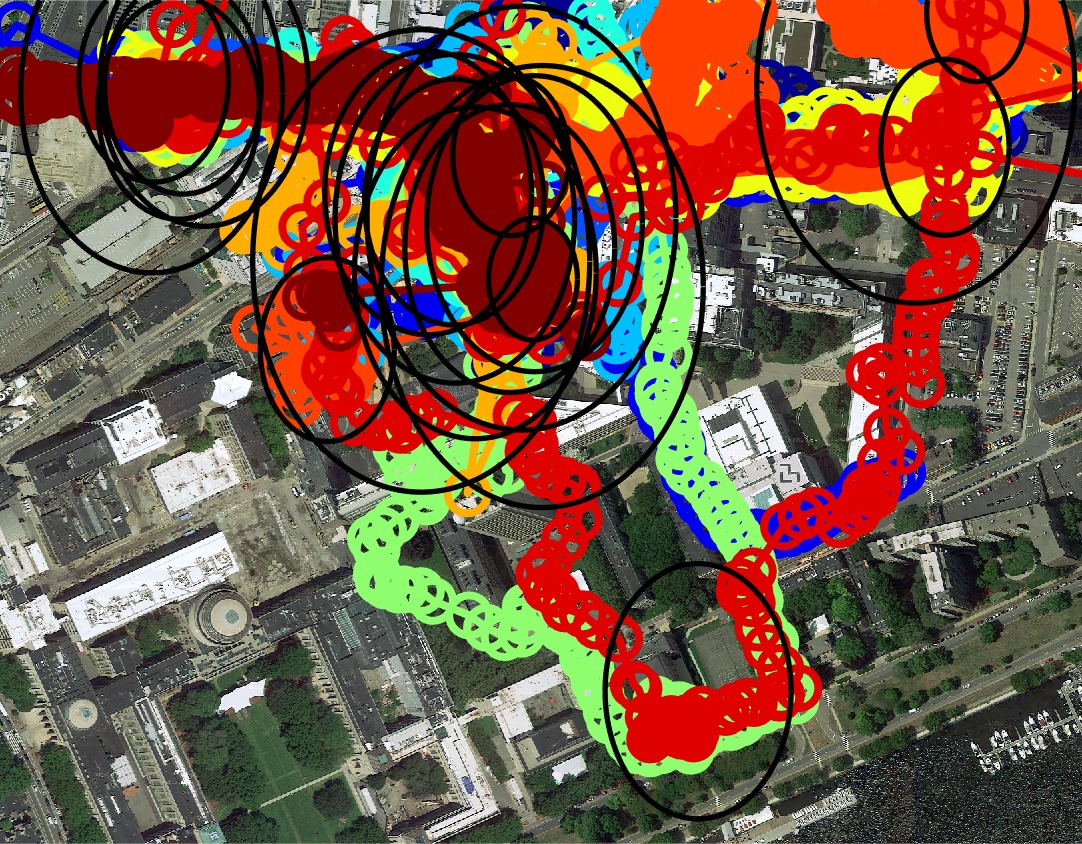}
\end{minipage}
\hfill
\begin{minipage}{0.45\linewidth}
\includegraphics[width=1\linewidth]{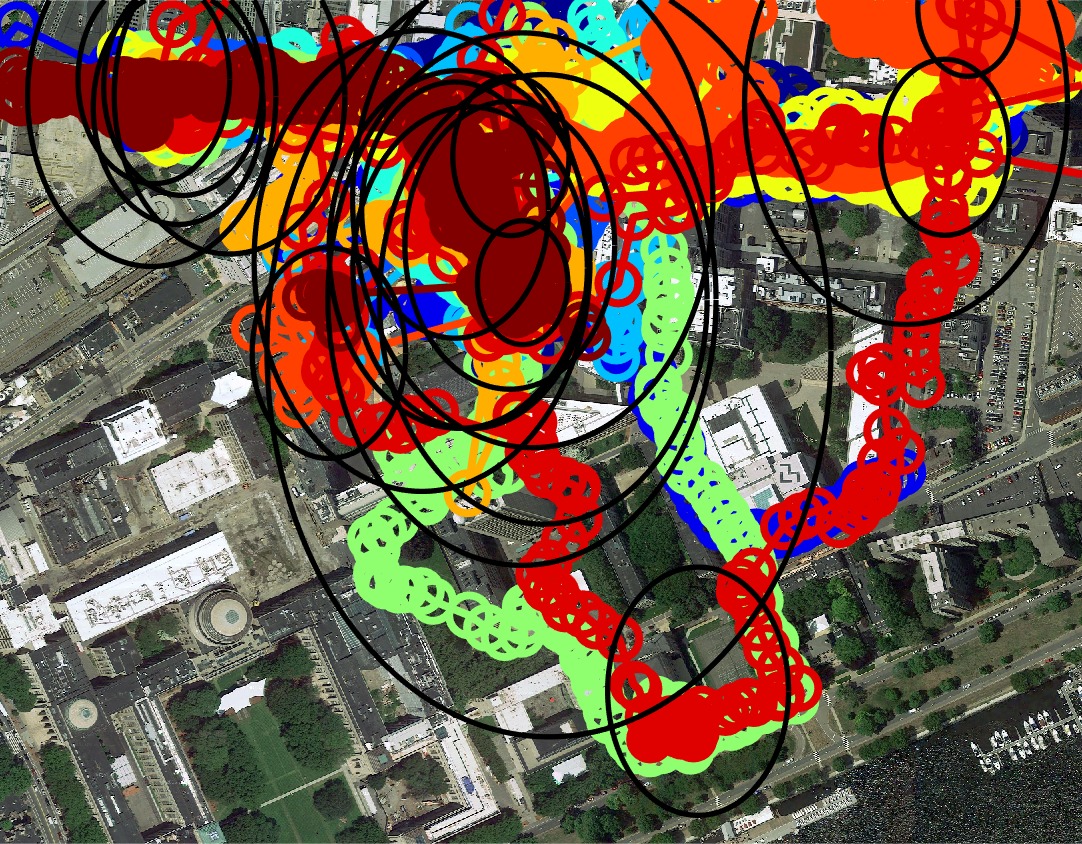}
\end{minipage}

\begin{minipage}{0.45\linewidth}
\includegraphics[width=1\linewidth]{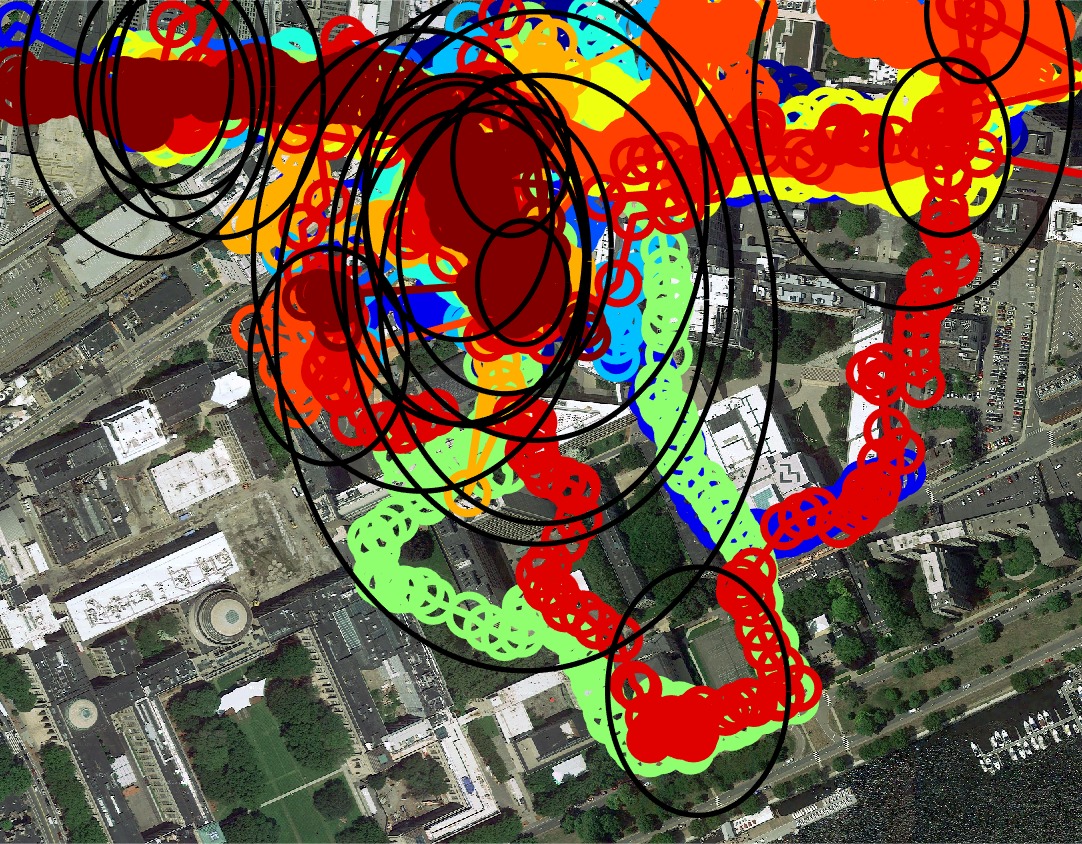}
\end{minipage}
\hfill
\begin{minipage}{0.45\linewidth}
\includegraphics[width=1\linewidth]{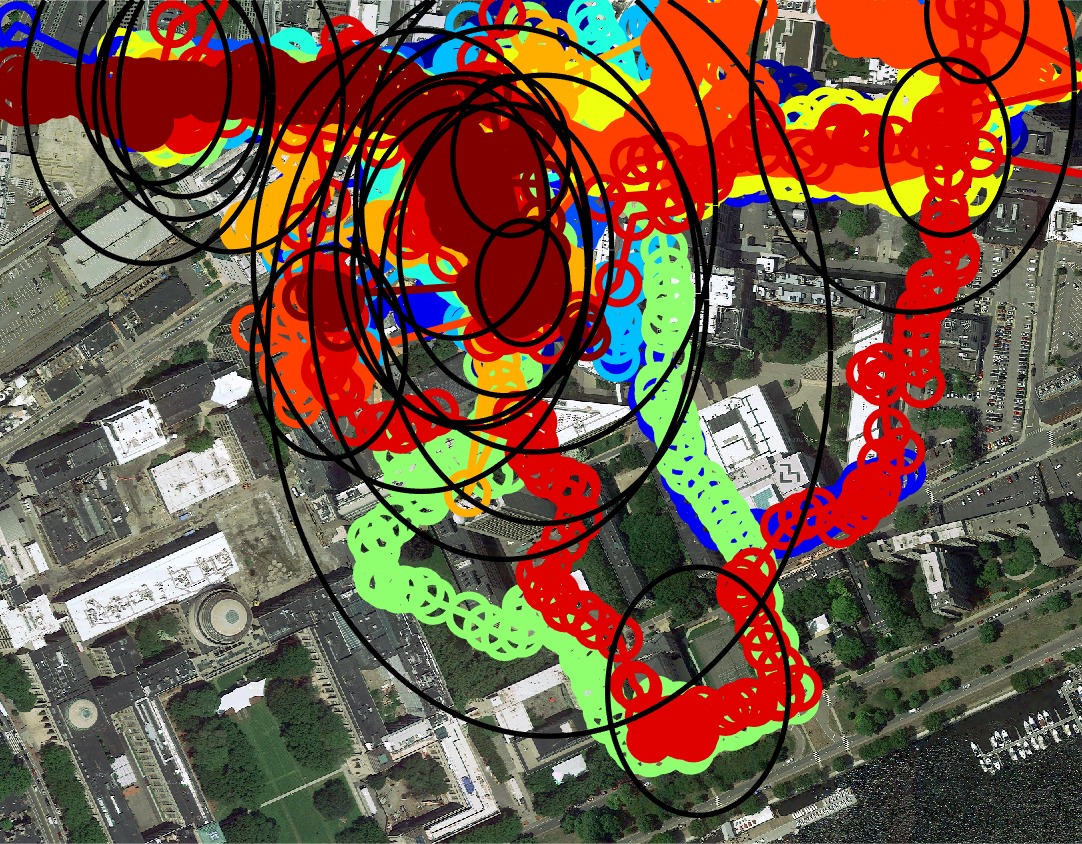}
\end{minipage}

\vspace{0.25cm}
\begin{minipage}{0.45\linewidth}
\includegraphics[width=1\linewidth]{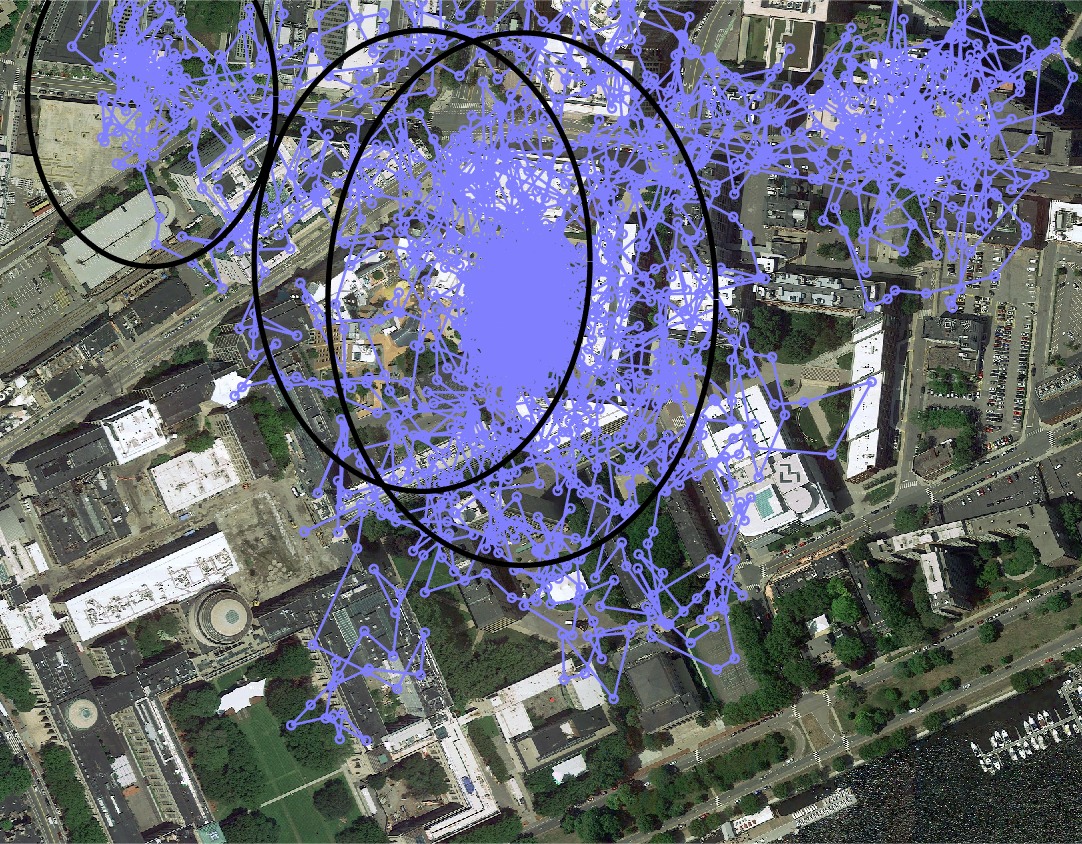}
\end{minipage}
\hfill
\begin{minipage}{0.45\linewidth}
\includegraphics[width=1\linewidth]{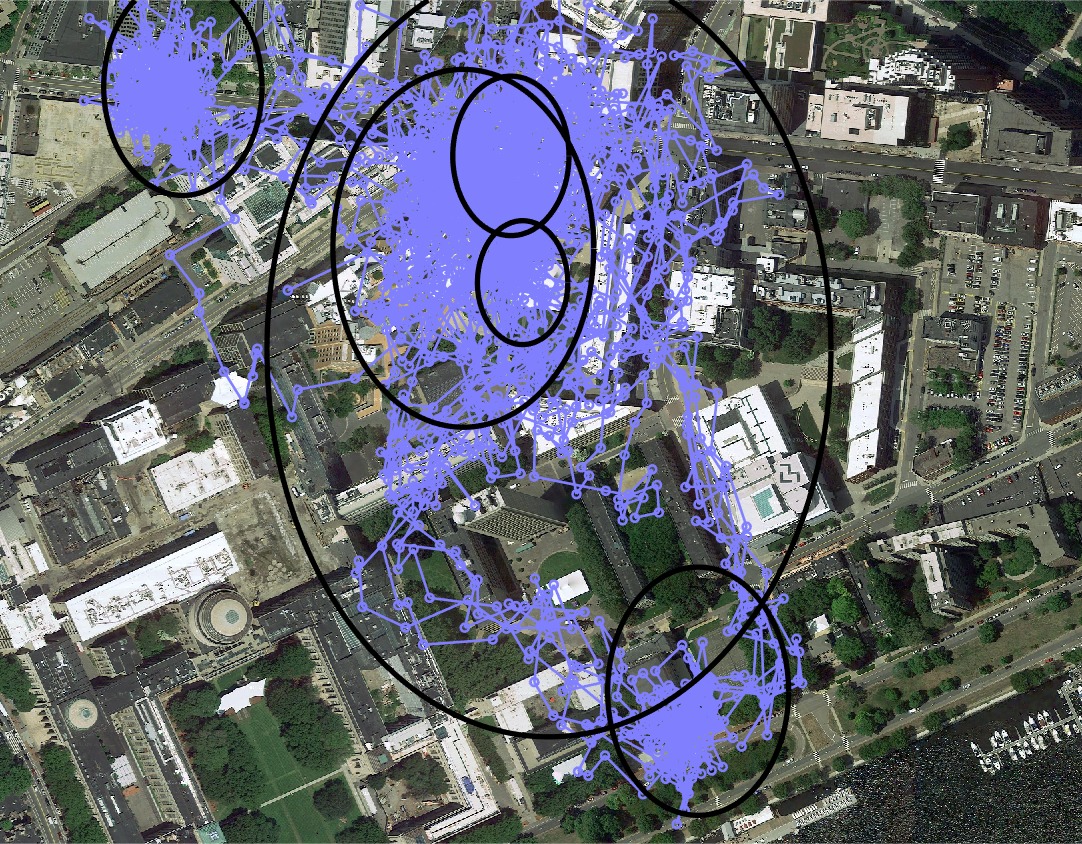}
\end{minipage}

\begin{minipage}{0.45\linewidth}
\includegraphics[width=1\linewidth]{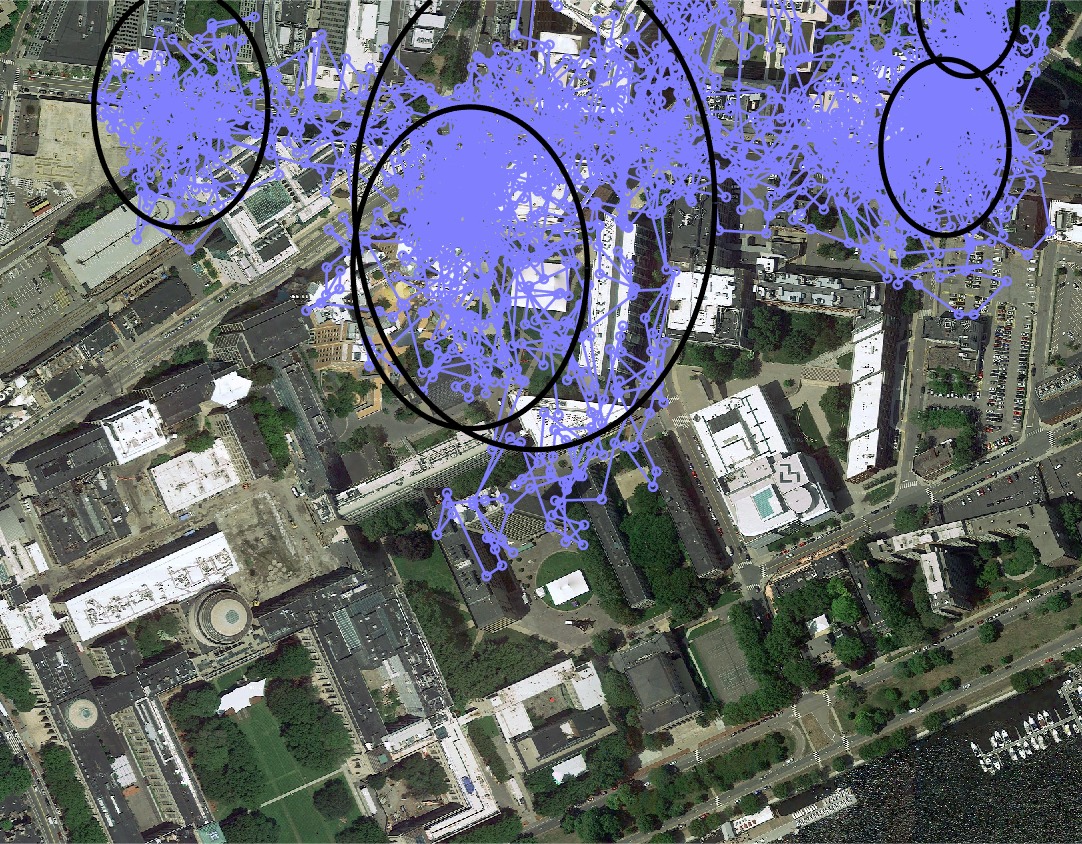}
\end{minipage}
\hfill
\begin{minipage}{0.45\linewidth}
\includegraphics[width=1\linewidth]{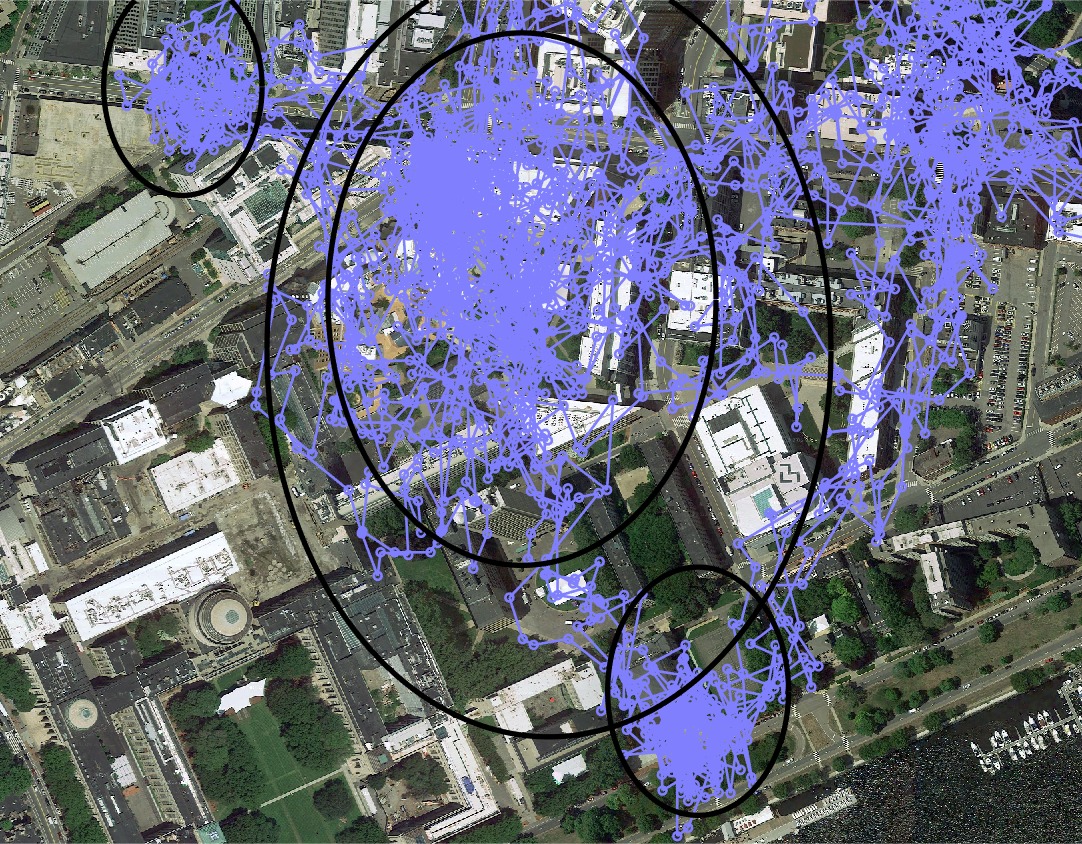}
\end{minipage}

\caption{Top two rows: example configurations sampled during the RJ-MCMC step (black circles), along with their corresponding participants' GPS measurements (multicolor line and circle). 
Bottom two rows: the activities related to four out of the 11 participants (black circles), and their estimated GP trajectories (light blue line and circle). \label{fig_sup_mat_conf} }
\end{figure}

\begin{figure}[htbp!]
\centering
\begin{minipage}{0.8\linewidth}
\includegraphics[width=1\linewidth]{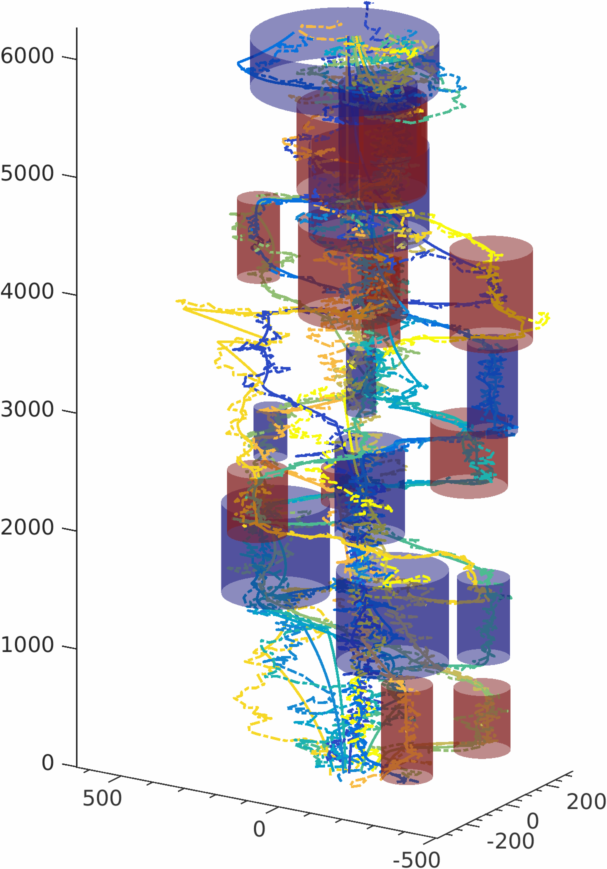}
\end{minipage}

\caption{Spatio-temporal plot of the maximum-probability configuration of activities. the vertical axis denotes time, the horizontal axes denote space. Cylinders denote detected activities. The paths denote samples of the GP trajectories of all 11 participants.
As can be seen, most meetings were detected in this configurations. \label{fig_sup_mat_spatiotemporal} }
\end{figure}

We now demonstrate the activities detected on a larger scale scenario. In this experiment, 11 participants were asked to walk in an urban area 800 meters across, and meet other participants for 3 or more minutes at each time, for 90 minutes (overall 1000 minutes, and more than 20 interactions).
The resulting detected activities, and the participants raw GPS trajectories that participated in each configuration are shown in Figure~\ref{fig_sup_mat_conf}. 
While not all configurations sampled contain all activities, we cover the set of expected activities and locations, as can be seen in the Figures~\ref{fig_sup_mat_conf},\ref{fig_sup_mat_spatiotemporal}. 
While the complexity of the scenario makes the visualization crowded, one can observe the scale of examples for which we can infer configurations, despite the limitation of MCMC methods in high-dimensions. 
We note that the locations and times match the locations reported by participants upto the localization artifacts expected in an urban area, with similar detection results as in the 4-participants example shown in the paper.

\section{Conclusions}
\label{sec:conclusions}
In this report we showed how a simple set of assumptions on social
interactions leads to a model that enables collaborative activity
summarization and improved localization under noisy and missing sensory streams. 
We demonstrate how a sampling-based strategy affords meaningful analysis even when the data is partial and noisy to the point that activities cannot be identified with certainty.
We expect additional observation models to allow extension of this framework to more refined activity classification and summarization, also for individual activities, as well as semantically-assisted GPS localization.

\section*{Appendix A: Collaborative Dynamical Models}
\label{sec:appendix1}
We now elaborate on the auxiliary observations mentioned in Subsection~2.3 of the paper.
Given a set of activities $\{A_i\}$, we model them as observations tying
participant locations to the activity. We propose two typical models,
though other models can also be proposed.

\textbf{Dynamic model} - in this model, we expect each participant to be at the
same location as all other participants at every time instance. The location
is not fixed and the group can move about. This is modeled by observations of
the form
\begin{align}
x_p(t)-x_p'(t) = n_{p,p',t}, n_{p,p',t} \sim  \mathcal{N}(0,\sigma^2)
\end{align}
for each participants pair $(p,p')$, time instance $t$, within the span of activity $A_i$.

\textbf{Static model} - in this model, we expect each participant to be at the same average location as are the other participants throughout the activity. We do not force the specific location, but do expect it to be fixed for the duration of the activity.

\begin{align}
x_p(t)-\sum_{p' \in A} \int_{t' \in span{A}} x_{p'}(t)dt = n_{p,t}, n_{p,t} \sim  \mathcal{N}(0,\sigma^2)
\end{align}
for each participant $p$ and time $t$ in the span of activity $A_i$.

We note that both of these terms are linear observations with additive white
Gaussian noise, and hence conditioning can be done by Gaussian belief
propagation \cite{citeulike:465818}.  Finally, in order to compute the
acceptance ratio, we compute the determinant of the conditioned covariance
matrix. This can be done with Cholesky factorization although faster
approximations have been proposed (see, for example,
\cite{DBLP:conf/icml/HanMS15}).

\section*{Appendix B: MCMC Steps}
\label{sec:appendix2}
We now describe the MCMC steps used to infer the configuration of
activities. Model inference uses a set of Gaussian
process samples: 500 samples for each actor, sparsely updated.  Our inference procedure includes
the following steps, illustrated in Table~\ref{tab_steps}.

\subsection*{Dimensionality Changing Steps}

\textbf{Birth} - In order to create a new activity, we first draw the number of participants based on a discretized log-Normal distribution
\begin{align}
 \hat{|p|} \sim \round{\log\mathcal{N}(2,0.5)},\ \ \ 
 |p|= max(\hat{|p|},2)
\end{align}
given the number of participants, we sample $|p|$ participants uniformly from all actors.
We sample the start time uniformly from the relevant measurements' support, and then the span. 
The prior distribution for the span is given as
\begin{align}
 \Delta t \sim \log\mathcal{N}(300,0.01),
\end{align}

\textbf{Death} - We propose death events (i.e removal of an activity from the configuration) by sampling uniformly from the existing
activity, and compute the probability of the remaining activities after removal
of one, for the acceptance probability.

\textbf{Split} - We split events along the time axis by uniformly sampling a
split point along the old span of the activity.

\textbf{Merge} - We merge two activities by taking the participants set and
spatial support of one of the activities, and merging the convexification of the
two events' temporal spans.  While a similar approach, of choosing a minimal
circle to include the two existing activities could be done for the spatial
support, we found it to be unnecessary in practice.

\setlength{\arrayrulewidth}{.075em}
\begin{table}
\centering
\begin{tabular}{|c|c|c|}

& Before & After\\
\hline
\begin{minipage}{0.22\linewidth}\centering\vspace{0.5em}
Birth
\vspace{0.5em}\end{minipage}
&
\begin{minipage}{0.22\linewidth}\centering\vspace{0.5em}
\includegraphics[height=0.5\linewidth]{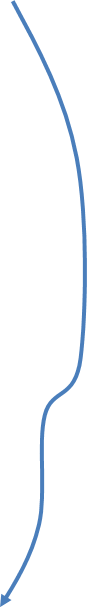}
\vspace{0.5em}\end{minipage}
&
\begin{minipage}{0.22\linewidth}\centering\vspace{0.5em}
\includegraphics[height=0.5\linewidth]{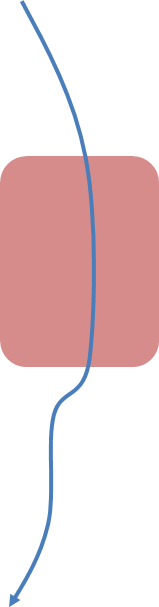}
\end{minipage}\\

\hline
\begin{minipage}{0.22\linewidth}\centering\vspace{0.5em}
Death
\vspace{0.5em}\end{minipage}
&
\begin{minipage}{0.22\linewidth}\centering\vspace{0.5em}
\includegraphics[height=0.5\linewidth]{Slide3.png}
\vspace{0.5em}\end{minipage}
&
\begin{minipage}{0.22\linewidth}\centering\vspace{0.5em}
\includegraphics[height=0.5\linewidth]{Slide2.png}
\end{minipage}\\

\hline
\begin{minipage}{0.22\linewidth}\centering\vspace{0.5em}
Split
\vspace{0.5em}\end{minipage}
&
\begin{minipage}{0.22\linewidth}\centering\vspace{0.5em}
\includegraphics[height=0.5\linewidth]{Slide3.png}
\vspace{0.5em}\end{minipage}
&
\begin{minipage}{0.22\linewidth}\centering\vspace{0.5em}
\includegraphics[height=0.5\linewidth]{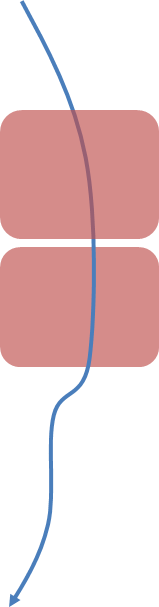}
\end{minipage}\\

\hline
\begin{minipage}{0.22\linewidth}\centering\vspace{0.5em}
Merge
\vspace{0.5em}\end{minipage}
&
\begin{minipage}{0.22\linewidth}\centering\vspace{0.5em}
\includegraphics[height=0.5\linewidth]{Slide8.png}
\vspace{0.5em}\end{minipage}
&
\begin{minipage}{0.22\linewidth}\centering\vspace{0.5em}
\includegraphics[height=0.5\linewidth]{Slide3.png}
\end{minipage}\\

\hline
\begin{minipage}{0.22\linewidth}\centering\vspace{0.5em}
Type
\vspace{0.5em}\end{minipage}
&
\begin{minipage}{0.22\linewidth}\centering\vspace{0.5em}
\includegraphics[height=0.5\linewidth]{Slide3.png}
\vspace{0.5em}\end{minipage}
&
\begin{minipage}{0.22\linewidth}\centering\vspace{0.5em}
\includegraphics[height=0.5\linewidth]{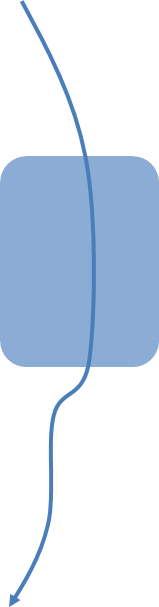}
\end{minipage}\\

\hline
\begin{minipage}{0.22\linewidth}\centering\vspace{0.5em}
Center
\vspace{0.5em}\end{minipage}
&
\begin{minipage}{0.22\linewidth}\centering\vspace{0.5em}
\includegraphics[height=0.5\linewidth]{Slide3.png}
\vspace{0.5em}\end{minipage}
&
\begin{minipage}{0.22\linewidth}\centering\vspace{0.5em}
\includegraphics[height=0.5\linewidth]{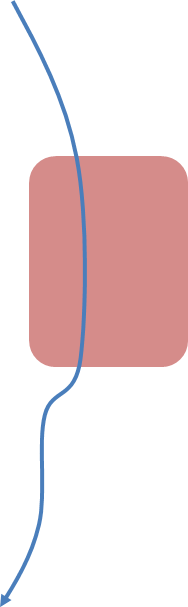}
\end{minipage}\\

\hline
\begin{minipage}{0.22\linewidth}\centering\vspace{0.5em}
Radius
\vspace{0.5em}\end{minipage}
&
\begin{minipage}{0.22\linewidth}\centering\vspace{0.5em}
\includegraphics[height=0.5\linewidth]{Slide3.png}
\vspace{0.5em}\end{minipage}
&
\begin{minipage}{0.22\linewidth}\centering\vspace{0.5em}
\includegraphics[height=0.5\linewidth]{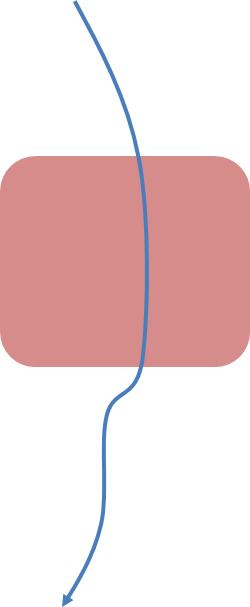}
\end{minipage}\\

\hline
\begin{minipage}{0.22\linewidth}\centering\vspace{0.5em}
Span
\vspace{0.5em}\end{minipage}
&
\begin{minipage}{0.22\linewidth}\centering\vspace{0.5em}
\includegraphics[height=0.5\linewidth]{Slide3.png}
\vspace{0.5em}\end{minipage}
&
\begin{minipage}{0.22\linewidth}\centering\vspace{0.5em}
\includegraphics[height=0.5\linewidth]{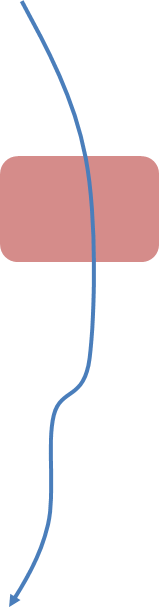}
\end{minipage}\\

\hline
\begin{minipage}{0.22\linewidth}\centering\vspace{0.5em}
Participants
\vspace{0.5em}\end{minipage}
&
\begin{minipage}{0.22\linewidth}\centering\vspace{0.5em}
\includegraphics[height=0.5\linewidth]{Slide3.png}
\vspace{0.5em}\end{minipage}
&
\begin{minipage}{0.22\linewidth}\centering\vspace{0.5em}
\includegraphics[height=0.5\linewidth]{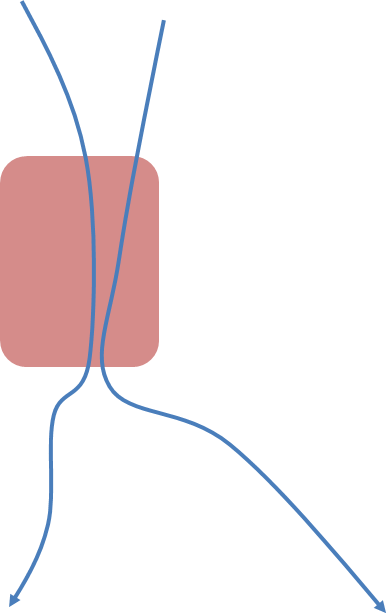}
\end{minipage}\\

\hline

\end{tabular}
 \caption{\label{tab_steps} Illustration of the MCMC steps used in the proposed activity detection algorithm. Blue curves represent participants trajectories, red and blue boxes portray two different types of activities.}
\end{table}

\subsection*{Parameter Changing Steps}
These are MCMC steps that do not modify the number of activities but rather
their attributes. They are: type, center, radius, span, start-time and
participants of activities. 

\textbf{Type} - We sample uniformly from all activity types, and compute the
proposed activities' probabilities conditioned on the sampled trajectories.

\textbf{Center} - We recompute the center of each activity by sampling with an
additive Gaussian displacement around the set of estimated GPS Gaussian process
samples. This limits our proposal distribution to be around the areas that are
relevant, assuming relatively mild GPS denial.

\textbf{Radius} - We use a proposal distribution for the radius,
\begin{align}
 q_r\sim \log\mathcal{N}(30,0.03).
\end{align}

\textbf{Span} - We sample the span according to a proposal distribution that
differs from its prior distribution in order to accomodate wider variations, as
supported by the data.  the proposal distribution is 
\begin{align}
 q_s \sim \log\mathcal{N}(300,0.005).
\end{align}

\textbf{Start-time} - we sample the start-time $t_i$ uniformly from the
temporal span of the measurements.

\textbf{Participants} - we resample participants according to the prior
distribution of $|p|$ and the uniform assumption for participants' identities.

\section*{Appendix C: Video Summarization}
\label{sec:appendix_video_summarization}
\newcommand{\cand}{\text{\emph{candidates}}}
We detail how to create a the video summarization as shown in the
supplementary video, based on the frames selected in Section~3.3 of the paper.
Considerations guiding the design of our video summarization include:
\begin{itemize}
  \item Variety \& Model Relevance: 
  \begin{itemize}
   \item Output frames should be balanced between desired actors, times and locations
   \item Output frames should emphasize inferred activities.

  \end{itemize}
\item Aesthetics: 
\begin{itemize}
\item Output frames should not be noisy, motion-blurred or
  of low content.
  \item Frames should not frequently/abruptly switch viewpoints/actors.
  \end{itemize}
\end{itemize}

\subsection*{Graph Formulation, Edge and Node Weights}
Summarization is formulated as an approximate, constrained shortest-path problem
on a trellis graph, similar to \cite{Arev:2014:AEF:2601097.2601198}, but
utilizing the sampled GPs, activities, and faces models from RJ-MCMC inference.
The trellis contains $m$ rows, one for each video stream, and has columns equal to the
maximum number of image frames associated with any actor. Each node, $V_i^t$,
corresponding to actor $i$'s possible output image at time $t$, is associated
with $f_i^t$, the candidate image frames of actor $i$ at output frame $t$. We
assume that if $|f_i^t|=0$, then it will never be chosen for time $t$ (since it
has no image to contribute). All nodes $V_i^t$ are connected to all nodes
$V_i^{t+1}$ by edges $E_{i,j}^t$, and both nodes and edges are weighted.

Nodes are weighted according to quality and level-of-activity metrics, edges
according to inter-frame distances with respect to sampled time and location,
activity and observed-participant continuity, as well as shared viewpoint. Node
weights are computed as cost $C(V_i^t) = C(f_i^t)$, where $f_i^t$ is actor $i$'s
candidate frame for time $t$:
\begin{equation} \label{node_cost}
  C(f_i^t) = \left( \left(1 - \frac{|{kp}_i^t|}{\overline{kp}}\right )w_q +
      \delta_{f}w_f + \delta_a w_a \right )w_i
\end{equation}
Where ${kp}_i^t$ are the SURF keypoints detected in frame $f_i^t$,
$\overline{kp} = \max_{i,t} |{kp}_i^t|$ is a normalizer, $\delta_a$ is an
indicator variable of whether a participant has been identified in
frame $f_i^t$, and $\delta_a$ is an indicator of whether $f_i^t$ \emph{belongs} to
an activity (i.e. $f_i^t$ is associated to actor $i$ who participates in an
activity, and $t$ is within that activity's span). Weights $w_q, w_f, w_a$
control relative importance and $w_i$ is a discount factor on the total cost for
actor $i$ (useful for favoring / excluding certain actors).

Although edge weights are computed between nodes $V_i^t$ and $V_j^{t+1}$, they
are in actuality functions of candidate frames $f_i^t$ and $f_j^{t+1}$, computed
as the distance:
\begin{align} \label{edge_cost}
  d(f_i^t, f_j^{t+1}) &= \left(
  \left( (1-\frac{M({kp}_i^t, {kp}_j^{t+1})}{\overline{M}({kp})}
  w_{\text{nm}} \right) + (1-\delta_{\text{sf}})w_{\text{sf}} +
  (1-\delta_{\text{sa}})w_{\text{sa}} \right.\nonumber\\
  &+ \qquad \left.\vphantom{\int_t}\Delta_{i,t}^{j,t+1}w_\Delta +
     \mathcal{T}_{i,t}^{j,t+1}w_\mathcal{T} \right)
  \delta_\mathcal{N}w_\mathcal{N}
\end{align}
Where $M({kp}_i^t, {kp}_j^{t+1})$ are the number of matched keypoints
between keypoints ${kp}_i^t$, ${kp}_j^{t+1}$ (defined above) and 
$\overline{M}({kp}) = \max_{i,t,j,t+1}M({kp}_i^t, {kp}_j^{t+1})$ is a
normalizer. $\delta_{\text{sf}}$ is an indicator for whether the same
participant is detected between frames $f_i^t, f_j^{t+1}$, and
$\delta_{\text{sa}}$ is an indicator for whether the two frames belong to the same
activity. $\Delta_{i,t}^{j,t+1}$ is the L2
spatial distance (marginalized from GP coordinates) between the location of actors $i$ and $j$
at the times when frames $k$ and $l$ were taken, and $\mathcal{T}_{i,t}^{j,t+1}$
is the L2 temporal distance between the times that frames $f_i^t$, $f_j^{t+1}$
were taken for actors $i, j$ at times $t, t+1$, respectively.
$\delta_\mathcal{N}$ is an indicator of whether frame $f_j^{t+1}$ belongs to a
new activity relative to previous frame selections,
and $w_\mathcal{N}$ is the discount factor applied in such a case.

\begin{figure}
\centering 
  \includegraphics[width=0.5\linewidth]{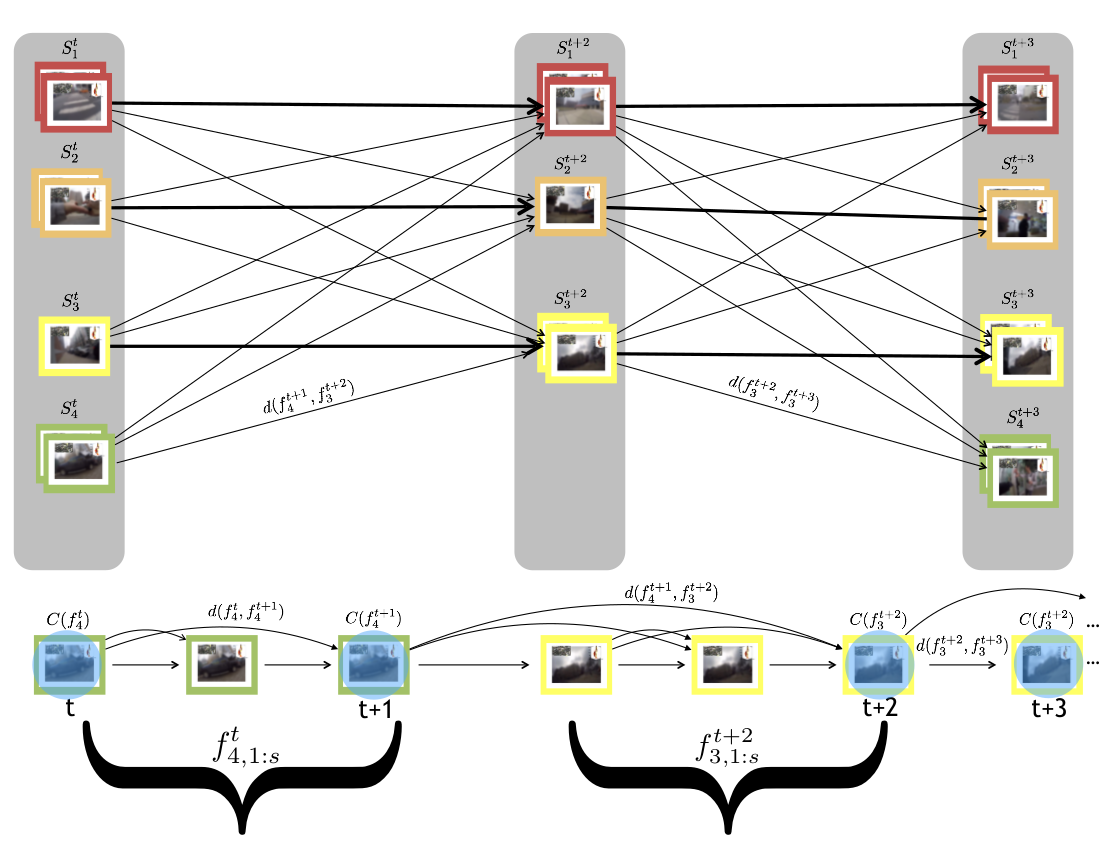}
  \caption{Illustration of the Super Trellis constrution used for video
    summarization. Rows correspond to actors (here, there are four), whereas 
    a column at time $t$ is responsible for anywhere from $1,
    \ldots, s$ output frames, where in this case $s=3$. Example Super Nodes are
    shown below the trellis, where $S_4^t$ contains frames $f_{4,1:s}^t$, and
    for which the first and third (translucent blue circles) were chosen as the
    lowest-cost output frames among all candidate Super Nodes $S_i^t$ for the
    first column. In this case, there is a toy constraint that the minimum and
    maximum number of consecutive output frames from a single actor are both
    $2$. Hence $S_4^{t+2}$ is not reachable in the second column (despite likely
    having frames associated with it) and so is omitted from the graph. Note
    also how at time $t+2$, only one frame was chosen from Super Node
    $S_3^{t+2}$, hence, by the constraint, $S_3^{t+3}$ must be chosen in the
    next column, though from that, only one node will be selected. There will
    typically be more observations, $f_{i,1:s}^t$ in Super Node $S_i^t$ than
    selected output frames. Note that here, only the edge and node costs that
    contributed to the actually-chosen path are displayed. }
  \label{fig:trellis}
\end{figure}

\subsection*{Constraint Handling}
Although the above costs empirically make good decisions about actor and
activity coverage, the resulting summarization includes rapid transitions
between actors' video feeds that are visually unpleasant for observers. To solve
this, and to increase control over the types of summaries produced, our
summarization supports the following constraints $c_1$: beginning/end time
$t_\text{begin}$, $t_{\text{end}}$, $c_2$: permitted/prohibited locations,
$c_3$: max temporal jump between successive output frames, and $c_4$: min/max
consecutive frames with respect to an actor. Though incorporation of these
constraints force an approximate, greedy solution rather than a true
shortest-path via Viterbi decoding, significantly more aesthetic videos are
produced as a result, and computation is faster.

\subsection*{Super Trellis and Super Node Construction}
To provide our algorithm the ability to "look ahead" (i.e. not make frame-wise
greedy decisions) and to ease the enforcement of constraint $c_4$, we pool nodes
along each row of the trellis into "Super Nodes" with pre-specified maximum
size of $s$ nodes. These Super Nodes can be said to form a "Super Trellis"
where, again, rows correspond to actors, but now each node corresponds to frames
$f_{i,1:s}$ for actor $i$. At each step through this Super Trellis the min-cost
path is computed within each Super Node, and then the min-cost Super Node is
chosen.  Hence, columns in the Super Trellis correspond to as many as $s$ time
steps in the original, frame-wise trellis. Node and edge costs in the Super
Trellis are handled as one would expect: edge weights between Super Nodes are
computed as the edge weights between the last frame of the previous Super Node
and the first frame of the next, and edges weights within Super Nodes are
computed as originally specified, though they are always between frames of the
same actor. The cost of each Super Node, then, is the sum of the internal node
and edge costs of the greedy-selected min-cost path within it (which, though it
always moves forward in time, need not select all frames). Details of this
algorithm are specified in Algorithm~(\ref{alg:video_summarize}).

\begin{algorithm}[ht!]
\caption{Collaborative Activity-Based Video Summarization \label{alg_summary}}
\label{alg:video_summarize}
\begin{algorithmic}[1]
  \STATE Given actors $a_{i=1:m}$, activity set $A_{1:n}$, actors'
    image sets $f_{i,:}$ with corresponding
    face recognitions $r_{i,:}$, desired number of output frames, $T$, as
    well as parameters $w_q, w_f, w_a, w_i, w_{\text{am}}, w_{\text{sf}},
    w_{\text{sa}}, w_\Delta, w_\mathcal{T}, w_\mathcal{N}$, super-node size $s$
    and constraints $c_1, \ldots, c_4$:
  \STATE $F \leftarrow [\quad], \text{curTime} \leftarrow t_\text{start},
         t \leftarrow 0$
  \STATE Precompute image keypoints for all $f_{i,:}$ as well as features
     matches between all candidate frames (restricted by $c_2, c_3$).
  \WHILE{ $|F| < N \text{ and curTime} \leq t_{\text{end}}$ }
    \STATE Compute Super Nodes $S_i^t$, each corresponding to actors' frames
      $f_{i,1:s}^t$ such that all frames meet constraints $c_{2:4}$ and have
      times at or beyond $\text{curTime}$. Note: $|f_{i,1:s}^t| \leq s$.
    \STATE $P \leftarrow [\quad]$
    \FOR{\textbf{each} $S \in S_i^t$}
    \STATE Greedy forward-select min-cost path according to node and edge
      weights (\ref{node_cost}, \ref{edge_cost}). Store in $P_i$.
    \ENDFOR
    \IF{ $|P_i| = 0, \forall i$ }
      \STATE break
    \ENDIF
    \STATE $F \leftarrow F \cup \hat{P}$ where $\hat{P}$ is the min-cost
      path (i.e. ordered set of image frames) chosen among paths $P_i$ in Super
      Nodes $S_i^t$.
    \STATE $\text{curTime} \leftarrow \text{time}(\hat{P})$ where
      $\text{time}$ is the aligned time of the final frame in path $\hat{P}$.
    \STATE $t \leftarrow t + |\hat{P}|$
  \ENDWHILE
  \STATE return $F$, the set of nodes indexing actors and frame numbers.
\end{algorithmic}
\end{algorithm}

\bibliographystyle{abbrv}
\bibliography{references}

\begin{thebibliography}{10}

\bibitem{adler2000}
R.~J. Adler.
\newblock On excursion sets, tube formulas and maxima of random fields.
\newblock {\em Ann. Appl. Probab.}, 10(1):1--74, 02 2000.

\bibitem{Arev:2014:AEF:2601097.2601198}
I.~Arev, H.~S. Park, Y.~Sheikh, J.~Hodgins, and A.~Shamir.
\newblock Automatic editing of footage from multiple social cameras.
\newblock {\em ACM Trans. on Graphics}, 33(4):81:1--81:11, July 2014.

\bibitem{bulling2014tutorial}
A.~Bulling, U.~Blanke, and B.~Schiele.
\newblock A tutorial on human activity recognition using body-worn inertial
  sensors.
\newblock {\em ACM Computing Surveys (CSUR)}, 46(3):33, 2014.

\bibitem{caba2015activitynet}
F.~Caba~Heilbron, V.~Escorcia, B.~Ghanem, and J.~Carlos~Niebles.
\newblock Activitynet: A large-scale video benchmark for human activity
  understanding.
\newblock In {\em CVPR}, pages 961--970, 2015.

\bibitem{DBLP:conf/cvpr/DaleSAP12}
K.~Dale, E.~Shechtman, S.~Avidan, and H.~Pfister.
\newblock Multi-video browsing and summarization.
\newblock In {\em CVPR Workshops}, pages 1--8, 2012.

\bibitem{Deisenroth_ITAC_2012}
M.~Deisenroth, R.~Turner, M.~Huber, U.~Hanebeck, and C.~Rasmussen.
\newblock Robust filtering and smoothing with {Gaussian} processes.
\newblock {\em IEEE Transactions on Automatic Control}, 2012.

\bibitem{DBLP:journals/corr/Fu14}
Y.~Fu.
\newblock Multi-view metric learning for multi-view video summarization.
\newblock {\em CoRR}, abs/1405.6434, 2014.

\bibitem{Fu:2010:MVS:2219116.2219978}
Y.~Fu, Y.~Guo, Y.~Zhu, F.~Liu, C.~Song, and Z.-H. Zhou.
\newblock Multi-view video summarization.
\newblock {\em Trans. Multi.}, 12(7):717--729, Nov. 2010.

\bibitem{Gonzalez85}
T.~F. Gonzalez.
\newblock Clustering to minimize the maximum intercluster distance.
\newblock {\em Theor. Comput. Sci.}, 38:293--306, 1985.

\bibitem{green1995reversible}
P.~J. Green.
\newblock Reversible jump {Markov} chain {Monte Carlo} computation and
  {Bayesian} model determination.
\newblock {\em Biometrika}, 82(4):711--732, 1995.

\bibitem{DBLP:conf/icml/HanMS15}
I.~Han, D.~Malioutov, and J.~Shin.
\newblock Large-scale log-determinant computation through stochastic chebyshev
  expansions.
\newblock In {\em Proceedings of the 32nd International Conference on Machine
  Learning, {ICML} 2015, Lille, France, 6-11 July 2015}, pages 908--917, 2015.

\bibitem{5589113}
J.~Hartikainen and S.~S\"{a}rkk\"{a}.
\newblock Kalman filtering and smoothing solutions to temporal gaussian process
  regression models.
\newblock In {\em Machine Learning for Signal Processing (MLSP)}, pages
  379--384, Aug 2010.

\bibitem{DBLP:conf/cvpr/HassnerHPE15}
T.~Hassner, S.~Harel, E.~Paz, and R.~Enbar.
\newblock Effective face frontalization in unconstrained images.
\newblock In {\em CVPR}, pages 4295--4304, 2015.

\bibitem{DBLP:conf/uai/HensmanFL13}
J.~Hensman, N.~Fusi, and N.~D. Lawrence.
\newblock {Gaussian} processes for big data.
\newblock In {\em UAI}, 2013.

\bibitem{Hochbaum85}
D.~Hochbaum and D.~Shmoys.
\newblock A best possible approximation for the k-center problem.
\newblock {\em Mathematics of Operations Research}, 10(2):180--184, 1985.

\bibitem{DBLP:conf/cvpr/HoshenBP14}
Y.~Hoshen, G.~Ben{-}Artzi, and S.~Peleg.
\newblock Wisdom of the crowd in egocentric video curation.
\newblock In {\em CVPR}, pages 587--593, 2014.

\bibitem{Huynh:2008:DAP:1409635.1409638}
T.~Huynh, M.~Fritz, and B.~Schiele.
\newblock Discovery of activity patterns using topic models.
\newblock In {\em UbiComp}, pages 10--19, New York, NY, USA, 2008. ACM.

\bibitem{joo_iccv_2015}
H.~Joo, H.~Liu, L.~Tan, L.~Gui, B.~Nabbe, I.~Matthews, T.~Kanade, S.~Nobuhara,
  and Y.~Sheikh.
\newblock Panoptic studio: A massively multiview system for social motion
  capture.
\newblock In {\em ICCV}, 2015.

\bibitem{KratzLeon10}
M.~F. Kratz and J.~R. Le\'{o}n.
\newblock Level curves crossings and applications for {G}aussian models.
\newblock {\em Extremes}, 13(3):315--351, 2010.

\bibitem{lee2015predicting}
Y.~J. Lee and K.~Grauman.
\newblock Predicting important objects for egocentric video summarization.
\newblock {\em Int. J. of Comp. Vision}, 114(1):38--55, 2015.

\bibitem{Liao:2007:EPA:1229555.1229562}
L.~Liao, D.~Fox, and H.~Kautz.
\newblock Extracting places and activities from {GPS} traces using hierarchical
  conditional random fields.
\newblock {\em IJRR}, 26(1):119--134, Jan. 2007.

\bibitem{Lu:2013:SSE:2514950.2516026}
Z.~Lu and K.~Grauman.
\newblock Story-driven summarization for egocentric video.
\newblock In {\em CVPR}, pages 2714--2721, 2013.

\bibitem{DBLP:conf/iccv/ParkJS13}
H.~S. Park, E.~Jain, and Y.~Sheikh.
\newblock Predicting primary gaze behavior using social saliency fields.
\newblock In {\em ICCV}, pages 3503--3510, 2013.

\bibitem{VideoDatasetReviewVersion}
{Removed for blind review}.
\newblock Collaborative video-{GPS} dataset, will be released with the paper.

\bibitem{shelley2005developing}
K.~J. Shelley.
\newblock Developing the american time use survey activity classification
  system.
\newblock {\em Monthly Lab. Rev.}, 128:3, 2005.

\bibitem{stein1999}
M.~L. Stein.
\newblock {\em {Statistical Interpolation of Spatial Data: Some Theory for
  Kriging}}.
\newblock Springer, 1999.

\bibitem{DBLP:conf/cvpr/TaigmanYRW14}
Y.~Taigman, M.~Yang, M.~Ranzato, and L.~Wolf.
\newblock Deepface: Closing the gap to human-level performance in face
  verification.
\newblock In {\em 2014 {IEEE} Conference on Computer Vision and Pattern
  Recognition, {CVPR} 2014, Columbus, OH, USA, June 23-28, 2014}, pages
  1701--1708, 2014.

\bibitem{TatJen06}
B.~Truong and S.~Venkatesh.
\newblock Utility-based summarization of home videos.
\newblock In T.-J. Cham, J.~Cai, C.~Dorai, D.~Rajan, T.-S. Chua, and L.-T.
  Chia, editors, {\em Advances in Multimedia Modeling}, volume 4351 of {\em
  LNCS}, pages 505--516. Springer Berlin Heidelberg, 2006.

\bibitem{citeulike:465818}
Y.~Weiss and W.~T. Freeman.
\newblock {Correctness of Belief Propagation in Gaussian Graphical Models of
  Arbitrary Topology}.
\newblock {\em Neural Computation}, 13(10):2173--2200, Oct. 2001.

\bibitem{weng2011event}
J.~Weng and B.-S. Lee.
\newblock Event detection in {T}witter.
\newblock {\em ICWSM}, 11:401--408, 2011.

\bibitem{DBLP:conf/iccv/WolfHT09}
L.~Wolf, T.~Hassner, and Y.~Taigman.
\newblock The one-shot similarity kernel.
\newblock In {\em ICCV}, pages 897--902, 2009.

\bibitem{yan2015egocentric}
Y.~Yan, E.~Ricci, G.~Liu, and N.~Sebe.
\newblock Egocentric daily activity recognition via multitask clustering.
\newblock {\em IEEE Trans. Image Process.}, 24(10):2984--2995, 2015.

\bibitem{6197723}
J.~Yang, J.~Luo, J.~Yu, and T.~Huang.
\newblock Photo stream alignment and summarization for collaborative photo
  collection and sharing.
\newblock {\em Multimedia, IEEE Transactions on}, 14(6):1642--1651, Dec 2012.

\bibitem{Yuan:2008:MGT:1460096.1460099}
J.~Yuan, J.~Luo, H.~Kautz, and Y.~Wu.
\newblock Mining {GPS} traces and visual words for event classification.
\newblock In {\em International Conference on Multimedia Information
  Retrieval}, MIR '08, pages 2--9, New York, NY, USA, 2008. ACM.

\bibitem{NAV:8292634}
P.~A. Zandbergen and S.~J. Barbeau.
\newblock Positional accuracy of assisted {GPS} data from high-sensitivity
  gps-enabled mobile phones.
\newblock {\em Journal of Navigation}, 64:381--399, 7 2011.

\bibitem{Zhang:2015:ASM:2716635.2659520}
Y.~Zhang, L.~Zhang, and R.~Zimmermann.
\newblock Aesthetics-guided summarization from multiple user generated videos.
\newblock {\em ACM Trans. Multimedia Comput. Commun. Appl.}, 11(2):24:1--24:23,
  Jan. 2015.

\bibitem{NIPS2014_5349}
B.~Zhou, A.~Lapedriza, J.~Xiao, A.~Torralba, and A.~Oliva.
\newblock Learning deep features for scene recognition using places database.
\newblock In {\em NIPS}, pages 487--495. 2014.

\end{thebibliography}

\end{document}